\documentclass[11pt]{article}

\usepackage[preprint]{acl}

\usepackage{times}
\usepackage{latexsym}
\usepackage[T1]{fontenc}
\usepackage[utf8]{inputenc}
\usepackage{microtype}
\usepackage{inconsolata}
\usepackage{soul}
\usepackage{graphicx}
\usepackage{tabularx}
\usepackage{listings}
\usepackage{wrapfig}
\usepackage{subcaption}
\usepackage{amsmath,amssymb,amsfonts,mathtools,amsthm}
\usepackage{booktabs,array,ragged2e,pdflscape,xltabular}
\usepackage{multirow,nicefrac,xcolor}
\usepackage[textsize=tiny]{todonotes}
\usepackage[most,skins,theorems]{tcolorbox}

\usepackage{amsmath,amsfonts,bm}

\def\eqref#1{equation~\ref{#1}}

\def\1{\bm{1}}

\DeclareMathAlphabet{\mathsfit}{\encodingdefault}{\sfdefault}{m}{sl}
\SetMathAlphabet{\mathsfit}{bold}{\encodingdefault}{\sfdefault}{bx}{n}

\hypersetup{
  pdftitle={SciGame: Let Agents Play the Scientific Discovery Games in the Real and Counterfactual World},
  pdfauthor={Yusheng Zhao, Bowei He, Xue Liu, Mingxiao Li}
}

\definecolor{failred}{RGB}{185,28,28}
\definecolor{amberx}{RGB}{176,102,0}

\newcolumntype{L}[1]{>{\raggedright\arraybackslash}p{#1}}

\tcbuselibrary{breakable}

\title{\textsc{SciGame}: Let Agents Play the Scientific Discovery Games in the Real and Counterfactual World}

\author{Yusheng Zhao$^{1}$, Bowei He$^{1}$,
Xue Liu${^{1,2}}$,
Mingxiao Li$^3$
  \\
         $^1$MBZUAI \quad $^2$McGill University\quad $^3$Shanghai AI Lab}

\begin{document}

\maketitle

\begin{abstract}
Large language model agents are increasingly used for scientific discovery. However, existing benchmarks mix law inference with programming, score data fitting instead of prediction on unseen inputs, and do not separate empirical evidence from pretrained priors. We introduce \textsc{SciGame}, a controlled benchmark that converts scientific literature into structured scientific cards. Each card yields a pair of Original and dimensionally consistent Counterfactual environments with the same task interface. \textsc{SciGame} contains a Historical collection of canonical laws and a live Temporal collection curated from papers published after model knowledge cutoffs. It evaluates agents in candidate-guided Identification and open-ended Discovery settings. Each setting runs under Prior-only, Fixed, and Adaptive protocols. Experiments on five frontier models yield three findings. First, familiar priors hurt performance once the governing relation is altered. Second, adaptive experimentation helps the strongest models but harms weaker ones under limited budgets. Third, good explanations can hide quantitative failures. Recovering the exact symbolic form remains the main bottleneck, even when agents interpret evidence correctly. Current agents can investigate systems actively, but inducing exact laws in unfamiliar settings remains an open challenge.
\end{abstract}

\section{Introduction}

Scientific discovery is a sequential process in which researchers formulate hypotheses, design experiments, interpret observations, and draw conclusions under specified conditions. LLMs increasingly support this process. They analyze scientific descriptions~\citep{m2024augmenting,laurent2024lab}, construct executable expressions~\citep{shojaee2025llmsrbench,lu2024ai}, interact with simulated environments~\citep{wang2022scienceworld}, and propose experiments to test hypotheses~\citep{boiko2023autonomous,huang2025popper,gottweis2026accelerating}.

Scientific law discovery requires more than fitting data~\citep{schmidt2009distilling,brunton2016discovering,udrescu2020ai}. An agent needs to infer how inputs determine observables and generalize to unqueried inputs~\citep{kronberger2022shape}. It needs to separate evidence from prior knowledge~\citep{kronberger2022shape,riosgarcia2026corral}, revise hypotheses that contradict observations~\citep{wilie2024belief,jhaveri2026failing}, and explain the relation and its applicability conditions. A benchmark should therefore evaluate the whole investigation~\citep{riosgarcia2026corral}.

Existing benchmarks cover parts of this process. Some evaluate experiment selection and hypothesis revision interactively~\citep{jansen2024discoveryworld,bertolazzi2026falsifybench,chen2025autobench}. Some recover equations from data~\citep{shojaee2025llmsrbench} or link modeling with active experimental design~\citep{gandhi2025boxinggym,kabra2026autoscience}. Others test reasoning under altered physical laws~\citep{koblischke2025gravity,chen2025physgym,zheng2026newtonbench,wiemann2026discoverphysics} or build tasks from publications~\citep{majumder2024discoverybench,chen2024scienceagentbench,starace2025paperbench,huang2026scilaws}.

\begin{table*}[t]
\centering
\caption{\textbf{Comparison of scientific law discovery benchmarks and our work.} ``$\checkmark$'' indicates supported and ``\textemdash{}'' unsupported. Paper grounding covers paper-derived tasks, Counterfactual includes mutated relation variants, and Temporal selection denotes date-based source curation.}
\label{tab:related_work_comparison}
\small
\setlength{\tabcolsep}{5pt}
\resizebox{\textwidth}{!}{%
\begin{tabular}{lccccc}
\toprule
Benchmark & \shortstack{Active\\experiments}
& \shortstack{Numerical\\prediction}
& \shortstack{Paper\\grounding}
& Counterfactual
& \shortstack{Temporal\\selection} \\
\midrule
DiscoveryWorld~\citep{jansen2024discoveryworld} & $\checkmark$ & $\checkmark$ & \textemdash & \textemdash & \textemdash \\
FalsifyBench~\citep{bertolazzi2026falsifybench} & $\checkmark$ & \textemdash & \textemdash & \textemdash & \textemdash \\
Auto-Discovery-Bench~\citep{chen2025autobench} & $\checkmark$ & $\checkmark$ & \textemdash & \textemdash & \textemdash \\
LLM-SRBench~\citep{shojaee2025llmsrbench} & \textemdash & $\checkmark$ & \textemdash & $\checkmark$ & \textemdash \\
BoxingGym~\citep{gandhi2025boxinggym} & $\checkmark$ & $\checkmark$ & $\checkmark$ & \textemdash & \textemdash \\
ActiveSciBench-Chem~\citep{kabra2026autoscience} & $\checkmark$ & $\checkmark$ & \textemdash & $\checkmark$ & \textemdash \\
Gravity-Bench~\citep{koblischke2025gravity} & $\checkmark$ & \textemdash & $\checkmark$ & $\checkmark$ & \textemdash \\
PhysGym~\citep{chen2025physgym} & $\checkmark$ & \textemdash & \textemdash & \textemdash & \textemdash \\
NewtonBench~\citep{zheng2026newtonbench} & $\checkmark$ & $\checkmark$ & \textemdash & $\checkmark$ & \textemdash \\
DiscoverPhysics~\citep{wiemann2026discoverphysics} & $\checkmark$ & $\checkmark$ & \textemdash & $\checkmark$ & \textemdash \\
DiscoveryBench~\citep{majumder2024discoverybench} & \textemdash & \textemdash & $\checkmark$ & \textemdash & \textemdash \\
ScienceAgentBench~\citep{chen2024scienceagentbench} & \textemdash & $\checkmark$ & $\checkmark$ & \textemdash & \textemdash \\
PaperBench~\citep{starace2025paperbench} & $\checkmark$ & $\checkmark$ & $\checkmark$ & \textemdash & $\checkmark$ \\
SciLaws-Bench~\citep{huang2026scilaws} & $\checkmark$ & $\checkmark$ & $\checkmark$ & $\checkmark$ & \textemdash \\
\midrule
SciGame (\textbf{Ours}) & $\checkmark$ & $\checkmark$ & $\checkmark$ & $\checkmark$ & $\checkmark$ \\
\bottomrule
\end{tabular}
}

\end{table*}

Four gaps remain. First, benchmarks often change the task, observation protocol, and metric together. Few compare prior-only reasoning, fixed observations, and adaptive experimentation on one interface, or pair original and counterfactual relations~\citep{chen2025physgym,zheng2026newtonbench,wiemann2026discoverphysics}. Second, common metrics may not test generalization to unseen inputs, and a plausible explanation can score well even when the predictor is wrong~\citep{riosgarcia2026corral}. Third, most benchmarks are static and rarely include relations published after a model's knowledge cutoff~\citep{white2025livebench,he2026livemathematicianbench}. Fourth, paper-based tasks mix law discovery with programming and replication~\citep{kon2026exp}.

We introduce \textsc{SciGame}, a controlled benchmark for scientific law discovery by LLM agents. It converts source relations into scientific cards that record the relation, variables, input domains, and applicability conditions. Each card yields two environments with the same interface. The \emph{Original} environment implements the source relation. The \emph{Counterfactual} environment applies a dimensionally consistent mutation. This pairing tests whether an agent follows evidence over familiar laws.

In \emph{Identification}, the agent selects the governing relation from candidates. In \emph{Discovery}, no candidates are given. In both, the agent submits executable expressions, an explanation, and applicability conditions. Each task runs under three protocols. Under \emph{Prior-only}, the agent receives no observations. Under \emph{Fixed}, it receives observations from a preset design. Under \emph{Adaptive}, it queries an \texttt{Oracle} for experiments and decides when to submit.

\textsc{SciGame} also includes a live, knowledge-cutoff-aware track. The Historical collection holds stable reference relations. The Temporal collection is built by a rerunnable pipeline that extracts relations from recent papers and reviews them independently. Its time window moves with model cutoffs to reduce memorization~\citep{rein2023gpqa,white2025livebench}.
We report a prediction score on hidden inputs, a hierarchical episode success rate, and a rubric-based explanation score~\citep{zheng2023judging,tan2025judgebench}.

Our contributions are as follows. First, we build a controlled benchmark with paired Original and Counterfactual environments and two task settings. Second, we introduce a live Temporal collection with an updatable curation pipeline. Third, we provide a unified framework that compares three protocols with three metrics. Finally, we evaluate multiple LLMs and find that familiar priors help on Original relations but hurt on Counterfactual ones. Adaptive experimentation helps stronger models but can harm weaker ones. Wrong functional relations are the main failure mode, and good explanations can coexist with incorrect predictors.

Code and data are available in this \href{https://anonymous.4open.science/r/scigame}{anonymous repository}.

\section{Relevant work}
\label{sec:background}

\label{sec:related_work}

\begin{figure*}[t]
\centering
\includegraphics[width=0.93\linewidth]{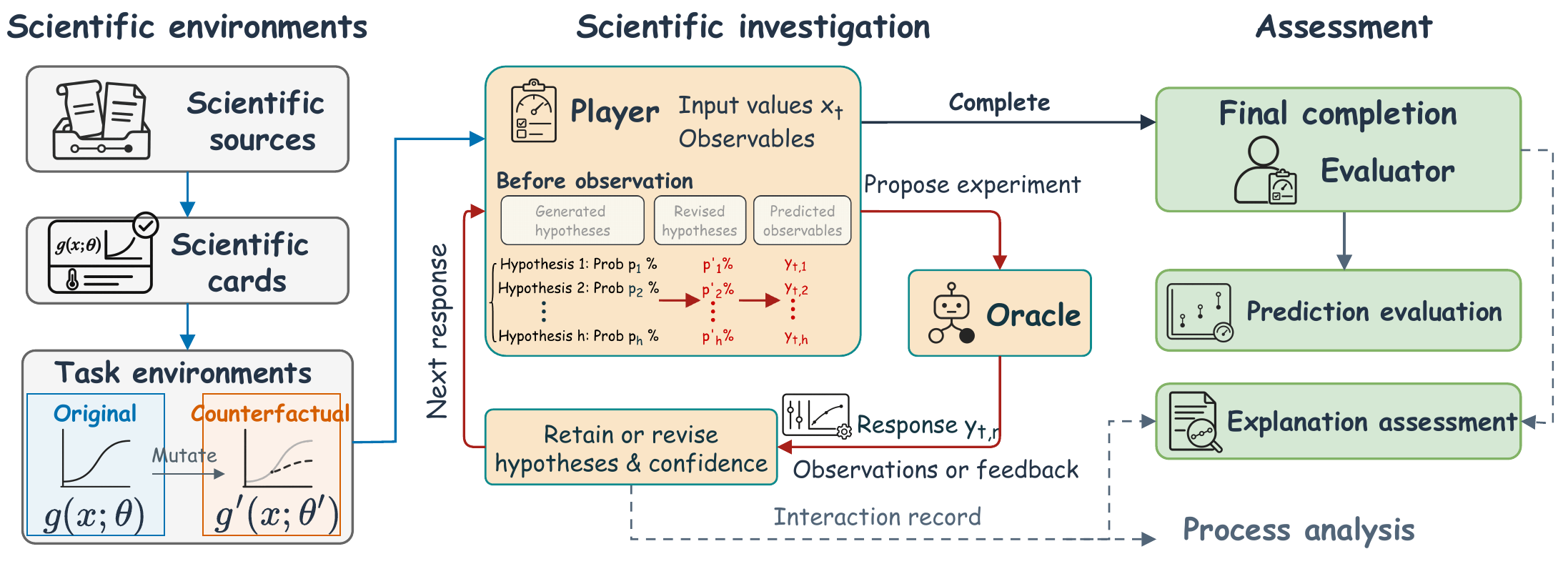}
\caption{\textbf{Overview of the study pipeline.} Scientific cards define the relations, variables, and applicability conditions used to build paired environments. The central panel illustrates the \textbf{Adaptive} protocol, where the \texttt{Player} designs experiments and chooses when to stop. Each episode is evaluated on predictions, explanations, and interaction trajectories.}
\label{fig:methodology_overview}
\end{figure*}

\textbf{Scientific discovery benchmarks.}
Scientific discovery benchmarks evaluate how agents infer relations and design experiments. DiscoveryWorld~\citep{jansen2024discoveryworld}, FalsifyBench~\citep{bertolazzi2026falsifybench}, and Auto-Discovery-Bench~\citep{chen2025autobench} study hypothesis revision and experiment selection. LLM-SRBench~\citep{shojaee2025llmsrbench}, BoxingGym~\citep{gandhi2025boxinggym}, and ActiveSciBench-Chem~\citep{kabra2026autoscience} evaluate equation discovery and active experimental design. Gravity-Bench~\citep{koblischke2025gravity}, PhysGym~\citep{chen2025physgym}, NewtonBench~\citep{zheng2026newtonbench}, and DiscoverPhysics~\citep{wiemann2026discoverphysics} vary the scientific context or underlying laws. None of them compares the same task with and without candidate expressions. It is therefore unclear how candidates affect prediction when observations support or contradict a reference relation.

\textbf{Paper-based scientific research.}
DiscoveryBench~\citep{majumder2024discoverybench} and ScienceAgentBench~\citep{chen2024scienceagentbench} assess hypothesis discovery and scientific programming. PaperBench~\citep{starace2025paperbench} evaluates research replication. These benchmarks measure broad research skills, not relation inference alone. SciLaws-Bench~\citep{huang2026scilaws} targets law discovery from papers, but its Real and Parallel settings use different observation and evaluation protocols. Their predictive accuracy is therefore not directly comparable.

In Table~\ref{tab:related_work_comparison}, we compare feature coverage of existing benchmarks and. \textsc{SciGame} covers all features. 

\section{Methodology}
\label{sec:methodology}

\subsection{Overview}
\label{sec:methodology_overview}
Figure~\ref{fig:methodology_overview} summarizes our evaluation pipeline and adaptive protocol. We distill scientific sources into structured scientific cards. Each card records a relation, parameter values, variable definitions, units, a valid input domain, and applicability conditions. Each card instantiates an Original environment and a paired Counterfactual environment. The two environments share the same task interface and experimental constraints. The Counterfactual environment applies a controlled structural mutation to the relation. An episode is one run of a \texttt{Player} on one environment. During an episode, the \texttt{Player} proposes experiments to an \texttt{Oracle} and receives observations or feedback. It updates its hypotheses and confidence values until it declares completion. We evaluate the final predictor and its scientific explanation, and we analyze the interaction trace to assess both task performance and investigation behavior. Sec.~\ref{sec:scientific_environments} describes source-relation curation and environment construction. Sec.~\ref{sec:experimental_protocol} details the experimental protocol, and Sec.~\ref{sec:evaluation} defines the evaluation metrics.

\textbf{Problem formulation.}
\label{sec:problem_formulation}
Each task specifies controllable inputs, observables, their units, and the conditions for running experiments. A deterministic simulator defines the response \(g_r(x)\) of observable \(r\) to an input vector \(x\) in the valid input domain. The relation and its parameters stay fixed within an episode. The \texttt{Player} builds a predictor \(\hat g\) with an executable expression \(\hat g_r(x)\) for each observable. It uses the task description, any candidate expressions, and the observations allowed by the protocol. It also writes a scientific explanation of the inferred relation and its applicability conditions.

\subsection{Data curation}
\label{sec:scientific_environments}

\textbf{Scientific data sources.}
We compile two collections of source relations. The \textbf{Historical} collection contains relations from the history of science. These relations serve as interpretable references. They test whether agents can use established relations to make predictions and support their conclusions with evidence. The \textbf{Temporal} collection extracts relations from publications in a configurable time window. We set this window after the reported knowledge cutoff of the evaluated \texttt{Player} to reduce data contamination.

\begin{figure*}[t]
\centering
\includegraphics[width=\linewidth]{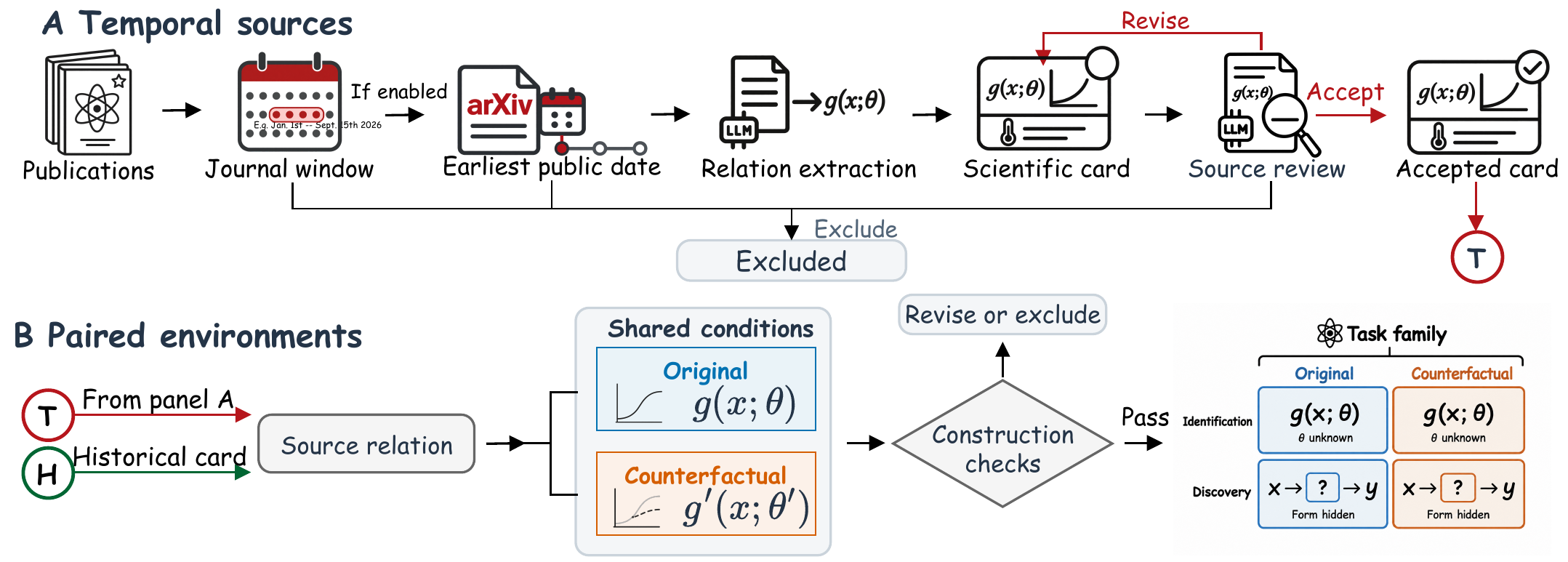}
\caption{\textbf{Data curation pipeline}. Part A describes the workflow on how to formulate accepted scientific cards from temporal sources. Part B describes how both historical and temporal corpus be curated as task family collections.}
\label{fig:data_curation}
\end{figure*}

\textbf{Scientific data construction.}
For each task family, we create a scientific card that records the relation, variables, input domains, and
applicability conditions. Historical relations use fixed modeling assumptions with explicit controllable inputs and observables. Figure~\ref{fig:data_curation} shows the live curation pipeline for the Temporal collection. It keeps only papers whose journal publication date and first preprint date both fall after the knowledge cutoff of the evaluated \texttt{Player}. An LLM extracts quantitative relations from these papers. A separate LLM reviewer checks each card for accuracy and completeness. Cards with missing or unsupported information are revised or removed. The pipeline can be rerun with new date windows to update the benchmark. Appendix~\ref{app:construction} lists all prompts.

\textbf{Counterfactual construction.}
For each task family in both collections, we pair an Original environment with a Counterfactual environment. The Original environment uses the source relation. The Counterfactual environment uses a systematically mutated version of it. Both environments share the same controllable inputs, input ranges, observables, and experimental constraints. They differ only in the relation that generates observations. This pairing tests whether the \texttt{Player} adapts to evidence that conflicts with established priors. Following NewtonBench~\citep{zheng2026newtonbench}, we apply structural mutations to coefficients and exponents, add nonlinear transformations, and introduce cross-input interactions. Nonlinear transformations keep dimensional consistency, and each mutated variable is updated in all related equations. Appendix~\ref{app:construction} describes the mutation protocols for Historical and Temporal families. Finally, we evaluate the ground-truth equations of each environment on observations from its counterpart. All of them fail the prediction evaluation, which confirms that the two environments can be told apart empirically.



\subsection{Experimental framework and protocol}
\label{sec:experimental_protocol}
We define two task settings for each Original and Counterfactual environment.

\begin{figure*}[t]
\centering
\includegraphics[page=1,width=\linewidth]{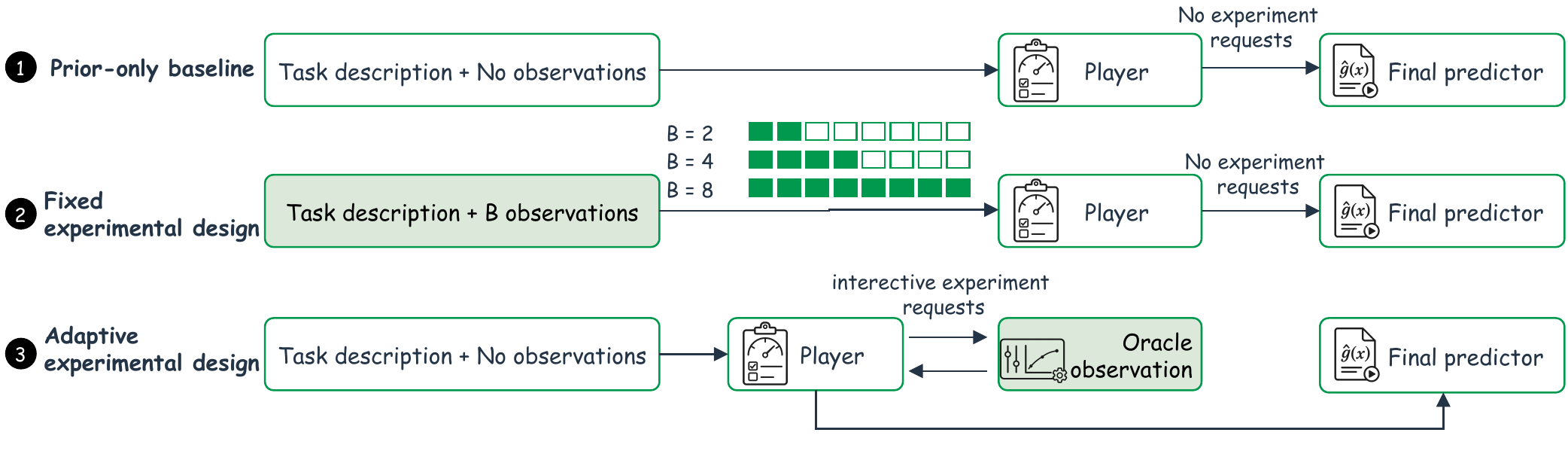}
\caption{\textbf{Comparison of observation protocols.} All three methods share the initial task description and assigned model information. \textbf{Fixed} provides a batch of observations from \(B\) pre-specified experiments upfront. \textbf{Adaptive} allows sequential experiment proposal without a pre-set budget constraint.}
\label{fig:experimental_protocols}
\end{figure*}

\textbf{Identification.} The \texttt{Player} receives a formula that combines the Original and Counterfactual expressions through an unknown parameter \(s\),
\begin{equation}
G_r(x;s)=(1-s)\, g_r(x)+s\, g'_r(x).
\end{equation}
Setting \(s=0\) gives the Original expression \(g_r(x)\), and \(s=1\) gives the Counterfactual expression \(g'_r(x)\). The task states whether intermediate values of \(s\) are allowed. In that case, the formula interpolates linearly between the two predictions. The simulator always sets \(s\) to zero or one, but this value is hidden from the \texttt{Player}.

\textbf{Discovery.} The \texttt{Player} receives no candidate expressions or parametric models. It builds a relation from inputs to observables using only the background description, experimental constraints, and available observations.

Each scientific family therefore yields four tasks. Both settings require executable expressions for the observables and a scientific explanation. Within an environment, they share the same experimental interface and numerical evaluation criteria.

An episode is one run of the \texttt{Player} on one task under one protocol. Each episode starts with an empty interaction history and is isolated from other episodes. Figure~\ref{fig:experimental_protocols} compares the three protocols.

(1) \textbf{Prior-only.} The \texttt{Player} receives no observations and does not interact with the \texttt{Oracle}. It completes the task using only the task description, protocol instructions, and prior knowledge.

(2) \textbf{Fixed.} The framework provides observations from a prespecified design with a fixed budget. The \texttt{Player} cannot select, modify, or repeat experiments. It then submits a final completion in the same format as Prior-only.

(3) \textbf{Adaptive.} The \texttt{Player} starts with no observations and proposes experiments to the \texttt{Oracle} step by step. At each LLM call, it receives the task description, protocol instructions, available observations, and interaction history. It returns a structured response with at least two hypotheses and their confidence scores, which are nonnegative and sum to one. The response also contains either an experimental proposal or a final completion. Each proposal includes a numerical prediction under every hypothesis for each proposed input and requested observable. Each final completion includes a predictor, a scientific rationale, and its applicability conditions.

The \texttt{Oracle} checks each proposal against the experimental constraints. It returns simulated observations for an accepted proposal. For a rejected proposal, it returns constraint feedback without any observations. The framework appends the proposal and the response to the interaction history. The \texttt{Player} then updates its hypotheses and either proposes another experiment or submits. The loop ends when the confidence of the top hypothesis exceeds a threshold \(\tau\).

\subsection{Evaluation metrics}
\label{sec:evaluation}
We use three metrics. The prediction score measures how accurately the submitted predictor generalizes to unseen inputs. The episode success rate measures task completion under numerical-tolerance and exact-expression criteria. The scientific explanation score evaluates how well the \texttt{Player} explains the relation, its consequences, and its applicability conditions. For each metric, we report the number of valid episodes and count invalid, missing, and undetermined episodes separately.

\begin{figure}[h]
    \centering
    \includegraphics[width=0.98\linewidth]{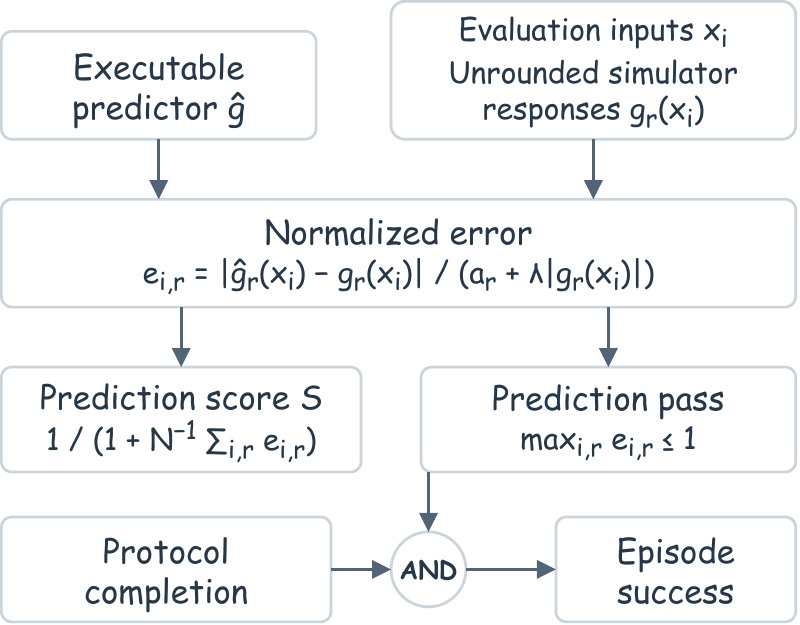}
    \caption{\textbf{Evaluation pipeline.}}
    \label{fig:protocols_evaluation}
\end{figure}

\textbf{Prediction score.}
The \texttt{Evaluator} compares the submitted predictions with the simulator responses on a hidden set of evaluation inputs. As shown in Figure~\ref{fig:protocols_evaluation}, we normalize the absolute errors by the declared tolerances and compute
\begin{equation}
\label{eq:prediction_score}
S=\frac{1}{1+{\frac{1}{N}\sum_{i,r} e_{i,r}}} \ \mbox{with} \  e_{i,r} = \frac{|\hat g_r(x_i)-g_r(x_i)|}{a_r+\lambda|g_r(x_i)|},
\end{equation}
where \(x_i\) is an evaluation input, \(a_r\) and \(\lambda\) are the absolute and relative tolerances, and \(N\) is the number of input and observable pairs. The score equals one at zero error and decreases as the error grows. It also separates completions that fail the tolerance but differ in accuracy.

\paragraph{Episode success rate.}
We use a hierarchical success standard. Under the \textbf{weak} standard, an episode succeeds if \(\max_{i,r} e_{i,r}\leq 1\) on the evaluation inputs and the completion follows the protocol. The \textbf{strict} standard also requires \(\hat g\) to match the ground-truth expression exactly. The success rate is \(\mathrm{ESR}=N_{\mathrm{s}}/N_{\mathrm{all}}\), where \(N_{\mathrm{s}}\) counts successful episodes and \(N_{\mathrm{all}}\) counts episodes with a determined outcome.

\textbf{Scientific explanation score.}
After all runs finish, an LLM judge scores each explanation post hoc using the task description, interaction history, and reference relation with its applicability conditions.
Three criteria assess whether the explanation correctly describes input--observable effects, gives a valid controlled comparison or scaling consequence, and stays within the supported evidence and assumptions; each is scored \(0\), \(0.5\), or \(1\), and the final score is their mean.
See Appendix~\ref{app:explanat} for the full rubric and procedure.

\section{Experiments}
\label{sec:experiments}

We evaluate the \texttt{Player} along two dimensions, varying the task setting (collection, environment, and whether candidate expressions are given) and the observation protocol (no observations, a fixed batch, or self-selected experiments), and report overall performance before using controlled comparisons to isolate the effects of prior knowledge, observations, and adaptive experimentation. More results are shown in Appendix~\ref{app:results}.

\subsection{Experimental setup}
\label{sec:experiments_setup}

  \begin{table*}[!t]
  \centering
  \caption{\textbf{Overall performance on both collections.} We report episode success rate (ESR, \%) under the \textbf{strict} standard, mean prediction score (PS) on hidden inputs, and mean scientific explanation score (SES). Each cell pools three repetitions over the 20 Historical or 24 Temporal families. Temporal papers were published between 30 April and 15 September 2026. Shaded cells are those referenced in the text.}
  \label{tab:experiments_main}
  \fontsize{8.5}{10.2}\selectfont
  \renewcommand{\arraystretch}{1.12}
  \setlength{\tabcolsep}{2pt}
  \resizebox{\linewidth}{!}{%
  \begin{tabular}{@{}ll rrr@{\hspace{5pt}}rrr@{\hspace{7pt}}rrr@{\hspace{5pt}}rrr@{\hspace{10pt}}rrr@{\hspace{5pt}}rrr@{\hspace{7pt}}rrr@{\hspace{5pt}}rrr@{}}
  \toprule
  & & \multicolumn{12}{c}{\textbf{Historical}} & \multicolumn{12}{c}{\textbf{Temporal}} \\
  \cmidrule(lr){3-14}\cmidrule(lr){15-26}
  & & \multicolumn{6}{c}{Identification} & \multicolumn{6}{c}{Discovery} & \multicolumn{6}{c}{Identification} & \multicolumn{6}{c}{Discovery} \\
  \cmidrule(lr){3-8}\cmidrule(lr){9-14}\cmidrule(lr){15-20}\cmidrule(lr){21-26}
  & & \multicolumn{3}{c}{Original} & \multicolumn{3}{c}{Counterfactual} & \multicolumn{3}{c}{Original} & \multicolumn{3}{c}{Counterfactual} & \multicolumn{3}{c}{Original} & \multicolumn{3}{c}{Counterfactual} & \multicolumn{3}{c}{Original} & \multicolumn{3}{c}{Counterfactual} \\
  \cmidrule(lr){3-5}\cmidrule(lr){6-8}\cmidrule(lr){9-11}\cmidrule(lr){12-14}\cmidrule(lr){15-17}\cmidrule(lr){18-20}\cmidrule(lr){21-23}\cmidrule(lr){24-26}
  Model & Method & ESR & PS & SES & ESR & PS & SES & ESR & PS & SES & ESR & PS & SES & ESR & PS & SES & ESR & PS & SES & ESR & PS & SES & ESR & PS & SES \\
  \midrule
  \multirow{3}{*}{\textbf{\texttt{DeepSeek-V4 Flash}}} & Prior-only & 88.3 & 0.950 & 0.707 & 0.0 & 0.136 & 0.401 & \hl{31.7} & 0.574 & 0.684 & \hl{0.0} & 0.100 & 0.218 & 40.3 & 0.467 & 0.563 & 0.0 & 0.158 & 0.552 & 0.0 & 0.078 & 0.439 & 0.0 & 0.062 & 0.379 \\
   & Fixed & 91.7 & 0.967 & 0.839 & 48.3 & 0.660 & 0.563 & \hl{53.3} & 0.867 & 0.757 & \hl{0.0} & 0.338 & 0.426 & 51.4 & 0.639 & 0.649 & 33.3 & 0.440 & 0.624 & 12.5 & 0.505 & 0.587 & \hl{9.7} & 0.378 & 0.564 \\
   & Adaptive & 76.7 & 0.819 & 0.799 & 33.3 & 0.456 & 0.601 & 23.3 & 0.485 & 0.713 & 0.0 & 0.186 & 0.488 & 52.8 & 0.597 & 0.697 & 44.4 & 0.536 & 0.723 & 9.7 & 0.406 & 0.578 & \hl{11.1} & \hl{0.270} & \hl{0.476} \\
  \midrule
  \multirow{3}{*}{\textbf{\texttt{K2-Horizon-375B}}} & Prior-only & 83.3 & 0.875 & 0.901 & 0.0 & 0.135 & 0.384 & \hl{25.0} & 0.450 & 0.817 & \hl{0.0} & 0.083 & 0.209 & 12.5 & 0.240 & 0.638 & 5.6 & 0.174 & 0.660 & 0.0 & 0.089 & 0.562 & 0.0 & 0.053 & 0.385 \\
   & Fixed & 78.3 & 0.842 & 0.969 & 36.7 & 0.424 & 0.802 & \hl{41.7} & 0.737 & 0.883 & \hl{0.0} & 0.218 & 0.527 & 40.3 & 0.453 & 0.851 & 27.8 & 0.350 & 0.818 & 13.9 & 0.351 & 0.717 & \hl{5.6} & 0.176 & 0.729 \\
   & Adaptive & 73.3 & 0.741 & 0.933 & 26.7 & 0.377 & 0.707 & 33.3 & 0.575 & 0.853 & 0.0 & 0.121 & 0.482 & 33.3 & 0.524 & 0.839 & 33.3 & 0.423 & 0.790 & 12.5 & 0.254 & 0.664 & \hl{4.2} & 0.104 & 0.570 \\
  \midrule
  \multirow{3}{*}{\textbf{\texttt{Qwen3.6 Plus}}} & Prior-only & 36.7 & 0.466 & 0.624 & 0.0 & 0.059 & 0.486 & \hl{13.3} & 0.208 & 0.645 & \hl{0.0} & 0.034 & 0.218 & 23.6 & 0.283 & 0.543 & 0.0 & 0.070 & 0.519 & 0.0 & 0.069 & 0.355 & 0.0 & 0.048 & 0.264 \\
   & Fixed & 36.7 & 0.448 & 0.868 & 1.7 & 0.090 & 0.385 & \hl{18.3} & 0.236 & 0.736 & \hl{0.0} & 0.047 & 0.240 & 25.0 & 0.312 & 0.580 & 13.9 & 0.186 & 0.527 & 12.5 & 0.263 & 0.539 & \hl{5.6} & 0.178 & 0.514 \\
   & Adaptive & 1.7 & 0.017 & 0.625 & 0.0 & 0.000 & 0.500 & 0.0 & 0.000 & 0.500 & 0.0 & 0.000 & 0.500 & 25.0 & 0.278 & 0.605 & 19.4 & 0.194 & 0.603 & 6.9 & 0.116 & 0.528 & \hl{2.8} & 0.050 & 0.504 \\
  \midrule
  \multirow{3}{*}{\textbf{\texttt{Qwen3.7 Plus}}} & Prior-only & 80.0 & 0.841 & 0.722 & 0.0 & 0.131 & 0.301 & \hl{31.7} & 0.560 & 0.667 & \hl{0.0} & 0.096 & 0.194 & 40.3 & 0.470 & 0.579 & 0.0 & 0.155 & 0.566 & 0.0 & 0.114 & 0.360 & 0.0 & 0.054 & 0.354 \\
   & Fixed & 81.7 & 0.870 & 0.900 & 38.3 & 0.463 & 0.634 & \hl{48.3} & 0.790 & 0.783 & \hl{0.0} & 0.208 & 0.434 & 40.3 & 0.523 & 0.693 & 33.3 & 0.372 & 0.706 & \hl{13.9} & 0.218 & 0.593 & \hl{6.9} & 0.145 & 0.586 \\
   & Adaptive & 46.7 & 0.502 & 0.708 & 15.0 & 0.170 & 0.576 & 21.7 & 0.366 & 0.670 & 0.0 & 0.079 & 0.462 & 34.7 & 0.408 & 0.616 & 27.8 & 0.312 & 0.664 & 6.9 & 0.114 & 0.569 & \hl{6.9} & 0.104 & 0.520 \\
  \midrule
  \multirow{3}{*}{\textbf{\texttt{GPT-5.5}}} & Prior-only & 73.3 & 0.780 & 0.884 & 31.7 & 0.404 & 0.519 & \hl{35.0} & 0.655 & 0.939 & \hl{0.0} & 0.123 & 0.289 & 8.3 & 0.244 & 0.517 & 19.4 & 0.389 & 0.768 & 0.0 & 0.209 & 0.587 & 0.0 & 0.081 & 0.530 \\
   & Fixed & 95.0 & 1.000 & 0.997 & 95.0 & 1.000 & 0.983 & \hl{63.3} & 0.998 & 0.943 & \hl{0.0} & 0.426 & 0.575 & 45.8 & 0.723 & 0.915 & 38.9 & 0.670 & 0.881 & 19.4 & 0.754 & 0.878 & \hl{9.7} & 0.422 & 0.715 \\
   & Adaptive & 98.3 & 1.000 & 1.000 & 88.3 & 0.966 & 0.992 & 63.3 & 0.981 & 0.947 & \hl{0.0} & 0.838 & 0.725 & 48.6 & 0.826 & 0.933 & 41.7 & 0.680 & 0.919 & 18.1 & 0.864 & 0.930 & 12.5 & 0.720 & 0.862 \\
  \midrule
  \multirow{3}{*}{\textbf{\texttt{GPT-5.6 Luna}}} & Prior-only & 83.3 & 0.871 & 0.933 & 1.7 & 0.133 & 0.397 & \hl{33.3} & 0.573 & 0.887 & \hl{0.0} & 0.122 & 0.247 & 11.1 & 0.216 & 0.611 & 0.0 & 0.117 & 0.485 & 0.0 & 0.185 & 0.519 & 0.0 & 0.077 & 0.408 \\
   & Fixed & 90.0 & 0.955 & 1.000 & 40.0 & 0.499 & 0.854 & \hl{55.0} & 0.928 & 0.871 & \hl{0.0} & 0.383 & 0.458 & 40.3 & 0.589 & 0.877 & 29.2 & 0.427 & 0.748 & 18.1 & 0.591 & 0.781 & \hl{11.1} & 0.397 & 0.699 \\
   & Adaptive & 83.3 & 0.884 & 0.953 & 28.3 & 0.365 & 0.860 & 45.0 & 0.765 & 0.881 & 0.0 & 0.299 & 0.601 & 47.2 & 0.637 & 0.843 & 26.4 & 0.336 & 0.822 & 11.1 & 0.370 & 0.821 & 8.3 & 0.250 & 0.805 \\
  \midrule
  \multirow{3}{*}{\textbf{\texttt{GPT-5.6 Terra}}} & Prior-only & 91.7 & 0.938 & 0.941 & 0.0 & 0.135 & 0.291 & \hl{26.7} & 0.575 & 0.914 & \hl{0.0} & 0.123 & 0.253 & 9.7 & 0.277 & 0.591 & 8.3 & 0.234 & 0.589 & 0.0 & 0.207 & 0.583 & 0.0 & 0.083 & 0.450 \\
   & Fixed & 93.3 & 1.000 & 1.000 & 85.0 & 0.952 & 0.947 & \hl{60.0} & 0.996 & 0.955 & \hl{0.0} & 0.391 & 0.497 & 41.7 & 0.653 & 0.858 & 38.9 & 0.582 & 0.815 & \hl{20.8} & 0.670 & 0.801 & \hl{8.3} & 0.426 & 0.635 \\
   & Adaptive & 91.7 & 0.950 & 0.986 & 73.3 & 0.807 & 0.954 & 45.0 & 0.748 & 0.980 & 0.0 & 0.341 & 0.658 & 55.6 & 0.706 & 0.891 & 41.7 & 0.670 & 0.857 & \hl{6.9} & 0.264 & 0.876 & \hl{1.4} & \hl{0.060} & \hl{0.815} \\
  \midrule
  \multirow{3}{*}{\textbf{\texttt{GPT-5.6 Sol}}} & Prior-only & 93.3 & 1.000 & 0.992 & 1.7 & 0.153 & 0.369 & \hl{36.7} & 0.656 & 0.942 & \hl{0.0} & 0.133 & 0.331 & 16.7 & 0.371 & 0.700 & 5.6 & 0.293 & 0.687 & 0.0 & 0.302 & 0.698 & 0.0 & 0.087 & 0.505 \\
   & Fixed & 98.3 & 1.000 & 1.000 & 96.7 & 1.000 & 0.997 & \hl{65.0} & 0.998 & 0.963 & \hl{0.0} & 0.449 & 0.647 & 55.6 & 0.739 & 0.901 & 45.8 & 0.693 & 0.880 & 20.8 & 0.752 & 0.850 & \hl{11.1} & 0.480 & 0.746 \\
   & Adaptive & 91.7 & 1.000 & 0.997 & 86.7 & 0.954 & 0.960 & 63.3 & 0.996 & 0.980 & 0.0 & 0.738 & 0.706 & 55.6 & 0.921 & 0.981 & 55.6 & 0.806 & 0.972 & 20.8 & 0.925 & 0.921 & 13.9 & 0.792 & 0.856 \\
  \midrule
  \multirow{3}{*}{\textbf{\texttt{GPT-6 Astra}}} & Prior-only & 78.3 & 0.992 & 0.924 & 0.0 & 0.137 & 0.425 & \hl{36.7} & 0.639 & 0.952 & \hl{0.0} & 0.130 & 0.417 & 8.3 & 0.438 & 0.710 & 0.0 & 0.259 & 0.659 & 0.0 & 0.319 & 0.690 & 0.0 & 0.089 & 0.576 \\
   & Fixed & 81.7 & 1.000 & 0.997 & 88.3 & 1.000 & 0.994 & \hl{66.7} & 0.999 & 0.994 & \hl{0.0} & 0.444 & 0.764 & 38.9 & 0.845 & 0.933 & 37.5 & 0.794 & 0.948 & 20.8 & 0.912 & 0.940 & \hl{9.7} & 0.527 & 0.819 \\
   & Adaptive & 80.0 & 1.000 & 1.000 & 85.0 & 1.000 & 1.000 & 63.3 & 0.999 & 1.000 & \hl{1.7} & 0.959 & 0.892 & 48.6 & 0.977 & 0.998 & 44.4 & 0.967 & 1.000 & 19.4 & 0.998 & 0.988 & \hl{16.7} & 0.995 & 0.988 \\
  \bottomrule
  \end{tabular}%
  }
\end{table*}

\textbf{Task collections.}
The Historical collection contains 20 task families based on classical scientific relations, listed in Table~\ref{tab:historical20}. The Temporal collection contains 24 task families from 23 papers published in leading peer-reviewed journals between 30 April and 15 September 2026, listed in Table~\ref{tab:temporal33}.


\textbf{\texttt{Player} models.}
We evaluate five models through Codex with \texttt{high} reasoning effort. They are \texttt{GPT-5.5}, \texttt{GPT-5.6 Luna}, \texttt{GPT-5.6 Terra}, \texttt{GPT-5.6 Sol}, and \texttt{GPT-6 Astra}. We also evaluate other advance open-sourced models such as \texttt{Qwen-3.6 Plus}, \texttt{Qwen-3.7 Plus}, \texttt{Deepseek-v4 Flash} and \texttt{K2-Horizon-375B-A23B} via APIs. Table~\ref{tab:player_models} lists the API release date and reported knowledge cutoff of each model. We interpret Temporal results relative to these cutoffs.

\textbf{Episodes and repetitions.}
We run three independent repetitions for each combination of task, model, and protocol. Each repetition is a new episode with an empty interaction history. No observations, responses, or records carry over between episodes.







\textbf{Evaluation procedure.}
We score each final predictor on hidden inputs with the prediction score and episode success criteria in Sec.~\ref{sec:evaluation}. We evaluate each scientific explanation separately with the rubric-based post hoc procedure. \texttt{GPT-6 Astra} serves as the fixed \texttt{Evaluator} for all \texttt{Player} models and conditions.

\subsection{Main benchmark results}
\label{sec:experiments_main}

\textbf{Overall performance.}
Table~\ref{tab:experiments_main} reports the episode success rate (ESR) and average prediction score (PS) of five \texttt{Player}s under the Prior-only, Fixed with \(B=4\), and Adaptive protocols. We summarize four findings.

(1) \textbf{Rooted priors act as a burden when learning counterfactual laws.}
In Original Discovery under Fixed, the average ESR is \(58.6\%\) on Historical tasks and \(21.3\%\) on Temporal tasks across all nine LLMs. This pattern reverses in Counterfactual Discovery. On Historical counterfactuals, strict ESR drops to zero for all LLMs. On Temporal counterfactuals, the average reaches \(8.9\%\). Under Adaptive, the top model also scores \(16.7\%\) on Temporal counterfactuals and \(1.7\%\) on Historical ones. This reversal suggests that familiar priors may keep LLMs from accepting counterfactual relations.

(2) \textbf{Adaptive interaction improves strong models but degrades weaker ones.}
For \texttt{GPT-6 Astra}, Adaptive raises the ESR on Temporal Counterfactual Discovery from \(9.7\%\) to \(16.7\%\); for \texttt{Qwen3.7 Plus} and \texttt{GPT-5.6 Terra}, it stays flat or drops, from \(6.9\%\) to \(6.9\%\) and from \(8.3\%\) to \(1.4\%\), respectively. On Temporal Original Discovery, \texttt{GPT-5.6 Terra} drops from \(20.8\%\) to \(6.9\%\) and \texttt{Qwen3.7 Plus} from \(13.9\%\) to \(6.9\%\). Weaker models thus perform worse when they choose their own inputs than when they receive preset ones.

(3) \textbf{Passive observations struggle on counterfactual laws despite succeeding on original ones. }
Under Fixed, the average ESR on Temporal Counterfactual Discovery is about half that on Temporal Original Discovery, at \(8.9\%\) and \(21.3\%\). All models fall in a narrow range from \(5.6\%\) to \(12.5\%\). Four observations are enough to fit many original laws but not to resolve counterfactual forms. Only Adaptive lifts the top model above this range.

(4) \textbf{High scientific explanations do not always guarantee predictive success.}
Among all combinations of model, protocol, and condition, several have a strict ESR below \(12.0\%\) and a scientific explanation score (SES) above \(0.700\); e.g., on Adaptive Temporal Counterfactual Discovery, \texttt{GPT-5.6 Terra} reaches an SES of \(0.815\) with an ESR of \(1.4\%\) and a PS of \(0.060\), and \texttt{DeepSeek-V4 Flash} reaches an SES of \(0.723\) with an ESR of \(11.1\%\). Neither metric alone describes an episode, so both are needed.

\begin{figure*}[ht]
\centering
\includegraphics[page=1,width=\linewidth]{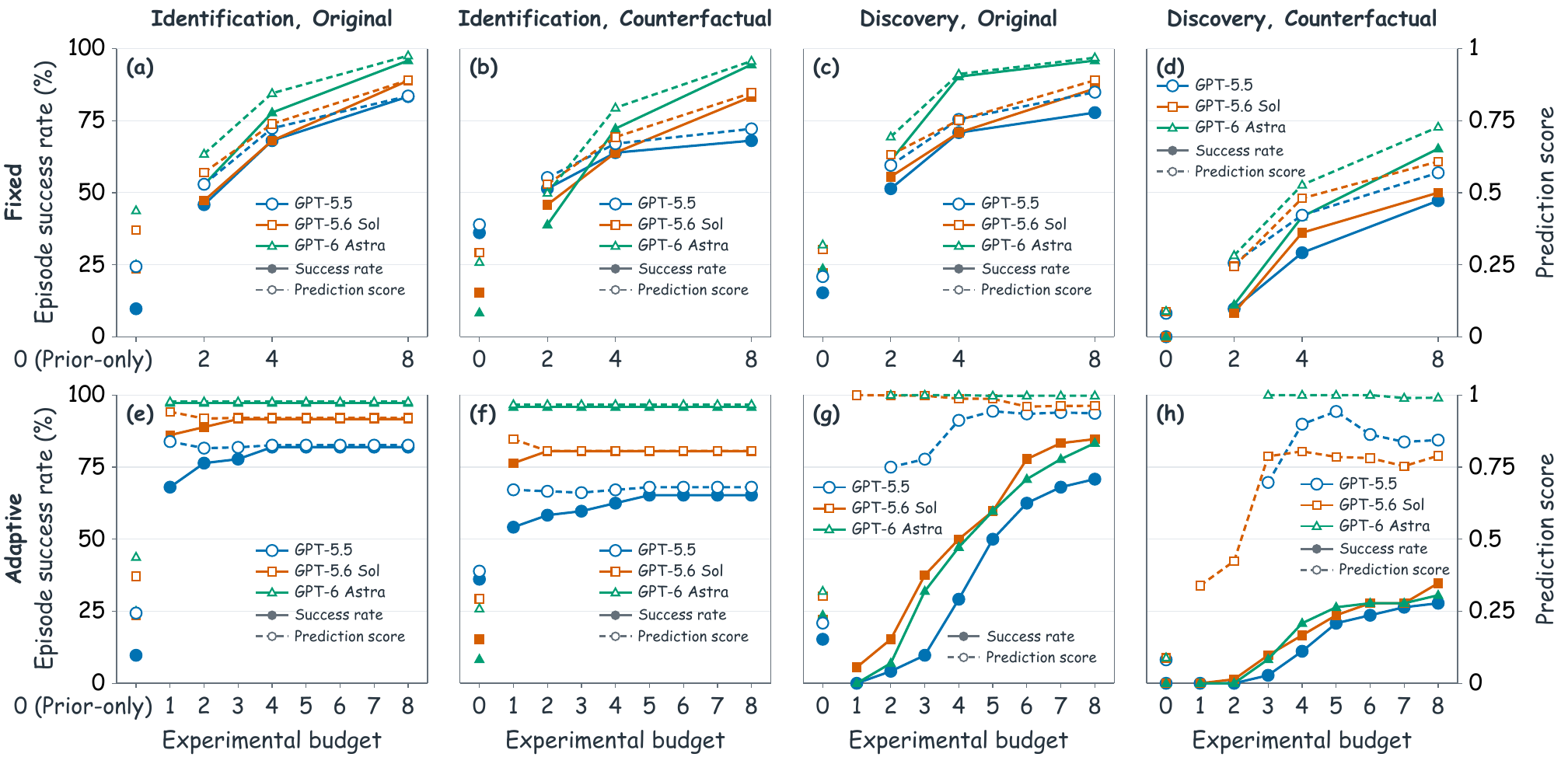}
\caption{\textbf{Performance of weak ESR on Temporal collection with varying experimental budget.} Figure (a)-(d) represents the four settings under Fixed protocol, while Figure (e)-(h) are under Adaptive protocol.}
\label{fig:temporal_budget}
\end{figure*}

\subsection{Experimental results and analysis}
\label{sec:experiments_ablations}

\textbf{Weak ESR on the Temporal collection under different budgets.}
Figure~\ref{fig:temporal_budget} shows how weak ESR changes with the observation budget \(B\) on the Temporal collection. For Adaptive, we report the cumulative success rate, which is the fraction of episodes solved within \(B\) steps. In Counterfactual Discovery at \(B=8\), Fixed reaches \(47.2\%\) to \(65.3\%\) across the three models, while Adaptive reaches only \(27.8\%\) to \(34.7\%\). For \texttt{GPT-6 Astra}, the gap is \(65.3\%\) under Fixed and \(30.6\%\) under Adaptive. Adaptive Discovery also has a clear cold start. Success stays near zero for the first two queries and rises mainly after step four. In contrast, Adaptive Identification saturates after one query. \texttt{GPT-6 Astra} solves over \(95\%\) of both Original and Counterfactual episodes at step one, and this rate stays flat up to \(B=8\).

Together with Table~\ref{tab:experiments_main}, these curves show that the advantage of Adaptive Discovery depends on a larger interaction budget. Figure~\ref{fig:budget_distribution} confirms this with the distribution of total budgets. With \(B\leq 8\), preset observations work better than sequential search for counterfactual laws. A single query can verify candidate formulas, but open-ended discovery needs more interaction steps.

\begin{figure}[h]
    \centering
    \includegraphics[width=\linewidth]{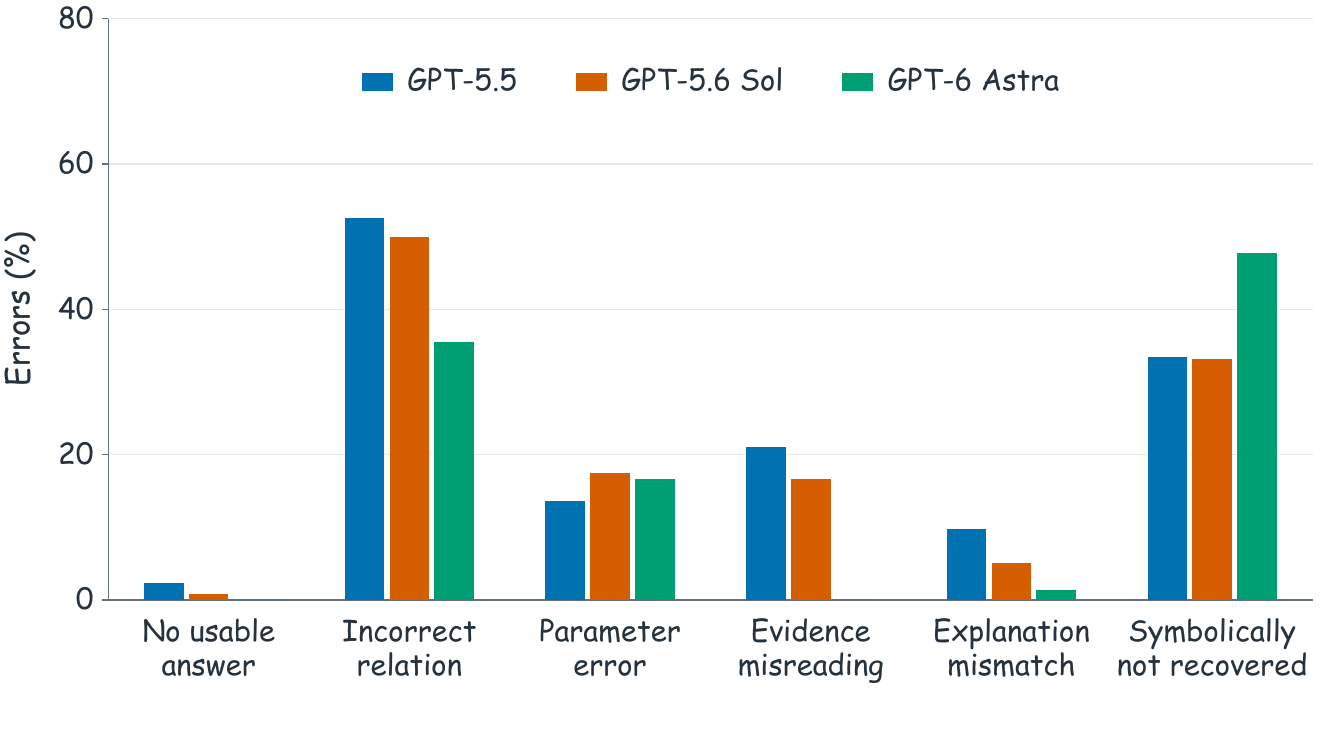}
    \caption{\textbf{Fail modes for GPT-5.5, GPT-5.6 Sol and GPT-6 Astra on two collection.}}
        \label{fig:fail_modes}
\end{figure}
\begin{figure}[h]
    \centering
    \includegraphics[width=0.9\linewidth]{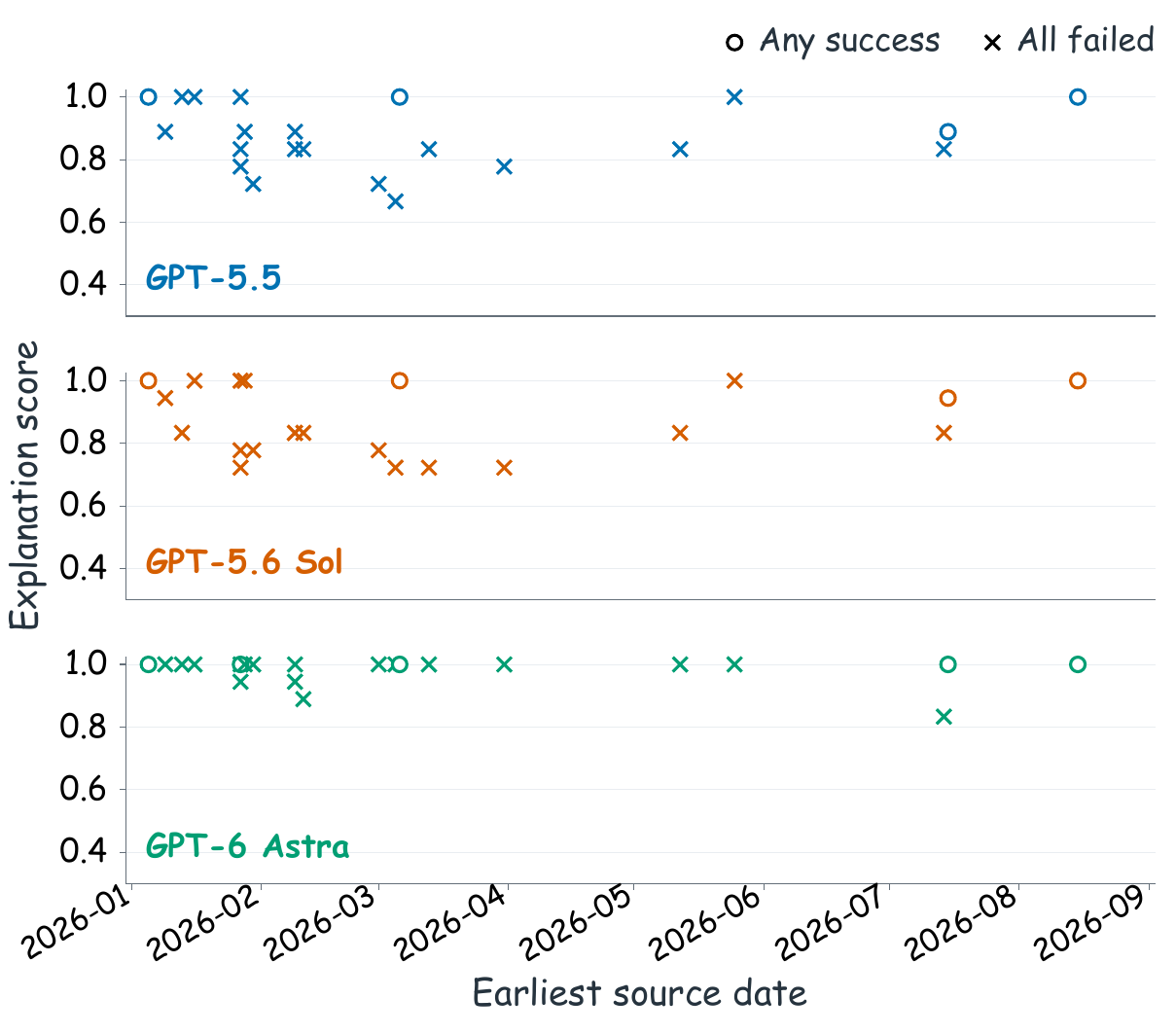}
    \caption{\textbf{Explanation Score for GPT-5.5, GPT-5.6 Sol and GPT-6 Astra on Discovery Counterfactual Temporal collection.}}
        \label{fig:explanation_scores}
\end{figure}

\textbf{Failure modes of the \texttt{Player}.}
We group errors in failed episodes into six categories: no usable answer, incorrect relation, parameter error, evidence misreading, explanation mismatch, and symbolically not recovered.
Figure~\ref{fig:fail_modes} shows the breakdown for \texttt{GPT} series models.
Incorrect relations account for about half of the failures for \texttt{GPT-5.5} and \texttt{GPT-5.6 Sol}, whereas \texttt{GPT-6 Astra} fails most often because \(48\%\) the exact equation is not symbolically recovered.
 Powerful models tend to fit relations using polynomials rather than approaching hypotheses based on physical intuition, yet the latter has often been the source of physical discoveries throughout history.

\textbf{Effect of data contamination.}
Figure~\ref{fig:explanation_scores} plots explanation scores and task outcomes in Temporal Counterfactual Discovery against each task family's earliest public release date. Circles~\({\Large \circ}\) denote families with at least one success in three repetitions, and crosses~\(\times\) denote families with no success. \texttt{GPT-6 Astra} solves more tasks from papers released up to September 2026 In contrast to \texttt{GPT-5.5} and \texttt{GPT-5.6 Sol}. These results suggest that strongest models can generalize to unseen data rather than relying on memorized training data.



\section{Conclusion}
In this paper, we presented \textsc{SciGame}, a controlled benchmark for scientific law discovery by LLM agents. It separates law inference from coding and other research tasks. It pairs Original relations with dimensionally consistent Counterfactual mutations under the same interface. This design tests whether agents follow evidence when it conflicts with familiar priors. \textsc{SciGame} compares the Prior-only, Fixed, and Adaptive protocols in both Identification and Discovery settings. Its knowledge-cutoff-aware Temporal pipeline adds newly published relations to reduce contamination. 

\newpage

\section*{Limitations}
Our experiments cover a small set of models, mostly from the GPT family, run through one agent harness. \textsc{SciGame} has 44 task families, most from physics, and each condition has three repetitions. Small differences between cells are therefore uncertain. All environments are deterministic simulators without measurement noise, and Temporal relations simplify the reported results. Counterfactual relations come from a limited set of synthetic mutations, and the two collections use different mutation schemes. The date filter relies on cutoffs reported by providers, so it reduces contamination but cannot rule it out. Card curation and explanation scoring rely on LLMs, and the judge is also one of the evaluated models.

\bibliography{references}

\newpage
\appendix

\section{Details of benchmark implementation}
\label{app:methodology}
This appendix specifies the Adaptive interaction protocol and the evalution pipeline. 


\subsection{Adaptive interaction protocol}
\label{app:adaptive_interaction}
\begin{figure}[h]
    \centering
    \includegraphics[width=\linewidth]{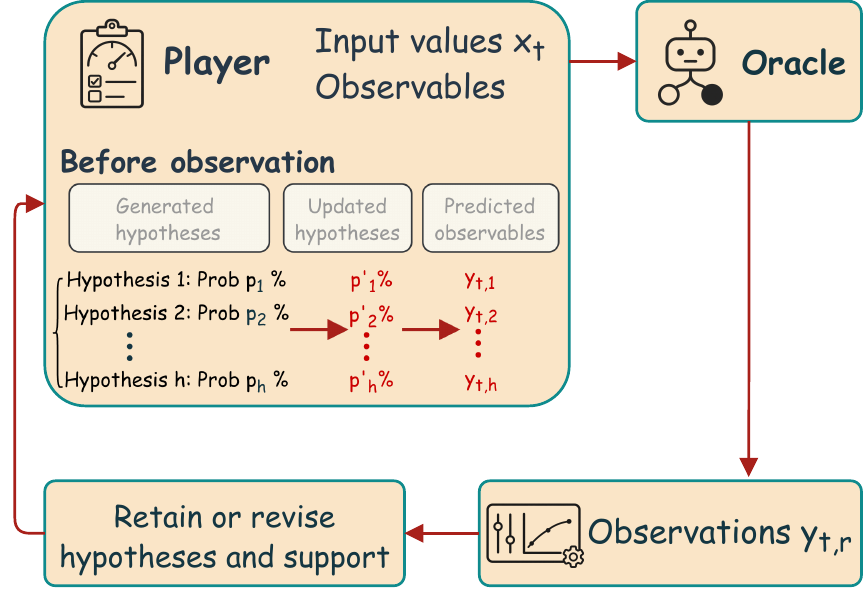}
    \caption{Adaptive interaction mechanism between the \texttt{Player} and the \texttt{Oracle}.}
    \label{fig:adaptive}
\end{figure}

Figure~\ref{fig:adaptive} illustrates the interaction loop under the Adaptive protocol. In each step, the \texttt{Player} submits a structured report bundling its current hypotheses, corresponding confidence scores, and an experimental proposal with prior expectation results. Upon receiving feedback from the \texttt{Oracle}, the \texttt{Player} either revises its hypotheses to design subsequent experiments or terminates the episode by submitting a final predictor under the stopping rule described below.

\textbf{Report schema and hypothesis tracking.}
Each hypothesis consists of a scientific statement and a reported confidence score in \([0,1]\). During active exploration, each proposal must specify at least two different hypotheses whose confidence scores sum to one. Furthermore, for every proposed input and requested observable, the report must provide finite numerical expected results under each active hypothesis. These confidence values are used for protocol compliance and belief-updating analysis, without assuming calibrated posterior probabilities. For the final completion (across Prior-only, Fixed, and Adaptive protocols), the submission must designate a selected hypothesis with an executable predictor \(\hat{g}\), a scientific explanation, and explicit applicability conditions. Table~\ref{tab:interaction_fields} summarizes the data schema across all interaction stages.

\textbf{Oracle validation and stopping criterion.}
Before executing any proposal, the \texttt{Oracle} validates the requested apparatus, input domains, observables, and experimental constraints. Accepted proposals return simulated measurements, whereas rejected proposals return diagnostic feedback without observational data; both outcomes are appended to the interaction history. Under the Adaptive protocol, the \texttt{Player} may request another experiment only while every reported hypothesis confidence remains strictly below the task threshold \(\tau\). Conversely, a valid final submission must select a hypothesis with confidence at least \(\tau\). Crucially, protocol compliance is evaluated independently of predictive accuracy: a submission violating the stopping rule can still be evaluated for its continuous prediction score \(S\) (see in Figure~\ref{fig:protocols_evaluation}), but the episode is automatically marked as a failure under the episode success rate (ESR).

\begin{table*}[h]
\centering
\caption{\textbf{Scientific fields recorded during an episode.} Each action by the \texttt{Player} contains a hypothesis report paired with either an experimental proposal or a final submission.}
\label{tab:interaction_fields}
\begin{tabularx}{\linewidth}{@{}p{0.25\linewidth}>{\raggedright\arraybackslash}X@{}}
\toprule
Record & Contents \\
\midrule
Hypothesis report & A statement, and confidence for each hypothesis. \\
Experimental proposal & Experimental objective, apparatus, input settings, requested observables, replicate count, and numerical predictions under each active hypothesis. \\
\texttt{Oracle} response & Execution status and feedback message. Accepted proposals include quantized measurements with units and resolution metadata. \\
Final submission & hypothesis, executable predictor expressions \(\hat{g}_r\) for all observables, scientific explanation, and applicability conditions. \\
\bottomrule
\end{tabularx}
\end{table*}

\paragraph{Deterministic numerical observations.}
For an accepted proposal at step \(t\), let \(x_t\) be the input vector and \(A_t\) the set of requested observable indices. The \texttt{Oracle} returns deterministic observations \(y_{t,r}\) defined as:
\begin{equation}
\label{eq:observation}
y_{t,r} = Q_r\!\left(g_r(x_t)\right), \qquad r \in A_t,
\end{equation}
where \(g_r(\cdot)\) is the ground-truth response function and \(Q_r\) quantizes the simulator response to the declared measurement resolution of observable \(r\). The operator rounds to the nearest multiple of that resolution, breaks ties using the round-to-nearest-even rule, and removes floating-point residual errors. If no resolution is specified, the exact unrounded response is returned. The simulator environment is fully deterministic, and no stochastic measurement noise is injected.

\textbf{Details of the three roles}
We evaluate five frontier GPT models as the \texttt{Player} through Codex, applying \texttt{high} reasoning effort across all experimental runs. Table~\ref{tab:player_models} summarizes the evaluated models alongside their access interface and reported knowledge cutoff dates. 

\begin{table*}[h]
    \centering
    \caption{The \texttt{Player} models and dates.}
    \label{tab:player_models}
    \small
    \begin{tabular}{@{}lll@{}}
        \toprule
        Player & Knowledge cutoff & API release \\
        \midrule
        GPT-5.5~\citep{openai2026gpt55}
            & 1 December 2025 & 24 April 2026 \\
        GPT-5.6 Luna~\citep{openai2026gpt56luna}
            & 16 February 2026 & 9 July 2026 \\
        GPT-5.6 Terra~\citep{openai2026gpt56terra}
            & 16 February 2026 & 9 July 2026 \\
        GPT-5.6 Sol~\citep{openai2026gpt56sol}
            & 16 February 2026 & 9 July 2026 \\
        GPT-6 Astra~\citep{openai2026gpt6astra}
            & 30 April 2026 & 3 September 2026 \\
        Qwen-3.6 Plus & February 2026 & 2 Apr 2026 \\
        Qwen-3.7 Plus & April 2026 & 2 June, 2026 \\
        DeepSeek-V4 Flash & Jul 2025 & 31 Jul 2026 \\
        {K2-Horizon-375B-A23B} & December 2025 & September 2026\\
        \bottomrule
    \end{tabular}
    
\end{table*}

The \texttt{Player} receives a system prompt governing action selection, hypothesis tracking, and pre-observation forecasting. This same prompt is also used in the Prior-only and Fixed protocols. The instruction prompt boxes retain the original wording, with omitted parts marked by \texttt{[...]}:
\begingroup
\tcbset{
  aibox/.style={
    width=\linewidth,
    top=8pt,
    bottom=4pt,
    colback=black!6!white,
    colframe=white!15!black,
    colbacktitle=black,
    enhanced,
    center,
    attach boxed title to top left={yshift=-0.1in,xshift=0.15in},
    boxed title style={boxrule=0pt,colframe=white,},
  },
  fontupper=\small,
  before upper={\raggedright\setlength{\parindent}{0pt}\setlength{\parskip}{0.45em}}
}
\newenvironment{ScigamePromptItems}[1]{%
  \begin{list}{}{%
    \setlength{\labelwidth}{#1}%
    \setlength{\labelsep}{0.6em}%
    \setlength{\leftmargin}{\dimexpr#1+0.6em\relax}%
    \setlength{\rightmargin}{0pt}%
    \setlength{\topsep}{0.2em}%
    \setlength{\itemsep}{0.35em}%
    \setlength{\parsep}{0pt}%
  }%
}{\end{list}}
\begin{tcolorbox}[aibox,breakable,title=Prompt for the \texttt{Player} instruction]
You are the \texttt{Player} in an interactive scientific modelling episode. Use only the public episode view and \texttt{Oracle} responses. Choose exactly one next action.

\texttt{[...]}

Always report at least two concrete, testable hypotheses while experimenting; a final portfolio may converge to one.

\texttt{[...]}

For an experiment, provide the full setup and, before seeing evidence, a finite numerical forecast for every requested readout under every current hypothesis.

\texttt{[...]}

In final.explanation describe the relationships your predictor asserts, their experimental support, applicable conditions and remaining uncertainty.

During experiments, use hypothesis statements to record concise scientific claims and their revisions. Explain which observations support retaining or changing a claim.

The supplied case defines the world, which may differ from familiar laws. Do not claim a mechanism beyond what the case and evidence establish.

\texttt{[...]}
\end{tcolorbox}
\endgroup

\subsection{Evaluation pipeline details}
\label{app:expression}

\textbf{Executable prediction representation}To isolate scientific law discovery from software engineering and code execution artifacts~\citep{chen2024scienceagentbench,starace2025paperbench}, each predictor \(\hat{g}\) that the \texttt{Player} completed is formulated as a pure arithmetic expression tree rather than arbitrary executable script code. In this representation, leaf nodes denote either finite numerical constants or references to controllable task inputs via \texttt{\{"var": "name"\}}. Internal nodes specify mathematical operations via \texttt{\{"op": "name", "args": [...]\}}, where each argument is itself a valid subexpression. Table~\ref{tab:expression_operators} lists all supported operators and their arities.

\begin{table*}[htbp]
\centering\small
\caption{Expression operators and their arities. Arity is the number of child
expressions.}
\label{tab:expression_operators}
\begin{tabularx}{\linewidth}{@{}>{\raggedright\arraybackslash}p{0.14\linewidth}>{\raggedright\arraybackslash}X@{}}
\toprule
Arity & Operators \\
\midrule
At least 2 & \texttt{add}, \texttt{multiply}, \texttt{maximum}, \texttt{minimum},
\texttt{hypot} \\
Exactly 2 & \texttt{subtract}, \texttt{divide}, \texttt{power}, \texttt{atan2} \\
Exactly 1 & \texttt{negate}, \texttt{exp}, \texttt{expm1}, \texttt{log},
\texttt{log1p}, \texttt{log2}, \texttt{log10}, \texttt{sqrt}, \texttt{cbrt},
\texttt{abs}, \texttt{sign}, \texttt{step}, \texttt{erf}, \texttt{erfc},
\texttt{sin}, \texttt{cos}, \texttt{tan}, \texttt{asin}, \texttt{acos},
\texttt{atan}, \texttt{sinh}, \texttt{cosh}, \texttt{tanh}, \texttt{asinh},
\texttt{acosh}, \texttt{atanh}, \texttt{floor}, \texttt{ceil}, \texttt{round} \\
\bottomrule
\end{tabularx}
\end{table*}
All operations follow standard real-domain semantics. Angles are evaluated in radians, and \texttt{atan2} accepts arguments in \((y, x)\) coordinate order. The operator \texttt{cbrt} evaluates the real cube root over \(\mathbb{R}\), \texttt{step(x)} yields one for \(x > 0\) and zero otherwise, and \texttt{sign(0)} evaluates to zero. Halfway rounding ties in \texttt{round} are broken using the round-to-nearest-even rule. All operations are strictly real-valued: divisions by zero, non-positive logarithmic arguments, complex outputs, and non-finite quantities (\(\mathrm{NaN}\) or \(\pm\infty\)) are strictly undefined and result in immediate invalidation of the predictor.

\textbf{Structural limits and validity criteria.}
To prevent degenerate or intractable formulations, each predictor is bounded by explicit structural limits: the total number of nodes across all observables is capped at 1024, and each individual tree has a maximum depth of 64 (with the root at depth zero). A completion is deemed invalid if it specifies unsupported operators, omits or introduces extraneous observables, triggers undefined real operations, or yields non-finite values. Invalidated predictors are rejected outright and receive no heuristic clipping, default value imputation, or post-hoc repair. Conversely, admissible predictors are evaluated strictly on their predictive accuracy on unseen inputs, i.e., the prediction score in \(S\) introduced in  Sec.~\ref{sec:evaluation}.

\textbf{Scientific explanation assessment.}
\label{app:explanat}
The scientific explanation score is assessed by an LLM \texttt{Evaluator} on the last \texttt{Player} action containing an active hypothesis, experimental rationale, final explanation, or applicability statement, conditioned on the complete interaction history. The \texttt{Evaluator} benchmarks correctness against the reference relation of the assigned environment, including the modified relation in a Counterfactual environment. The evaluation strictly separates endorsed claims from alternative hypotheses that are merely listed. The instruction prompt used for the LLM evaluator is as follows, with omissions marked by \texttt{[...]}:
\begingroup
\tcbset{
  aibox/.style={
    width=\linewidth,
    top=8pt,
    bottom=4pt,
    colback=black!6!white,
    colframe=white!15!black,
    colbacktitle=black,
    enhanced,
    center,
    attach boxed title to top left={yshift=-0.1in,xshift=0.15in},
    boxed title style={boxrule=0pt,colframe=white,},
  },
  fontupper=\small,
  before upper={\raggedright\setlength{\parindent}{0pt}\setlength{\parskip}{0.45em}}
}
\newenvironment{ScigamePromptItems}[1]{%
  \begin{list}{}{%
    \setlength{\labelwidth}{#1}%
    \setlength{\labelsep}{0.6em}%
    \setlength{\leftmargin}{\dimexpr#1+0.6em\relax}%
    \setlength{\rightmargin}{0pt}%
    \setlength{\topsep}{0.2em}%
    \setlength{\itemsep}{0.35em}%
    \setlength{\parsep}{0pt}%
  }%
}{\end{list}}
\begin{tcolorbox}[aibox,breakable,title=Prompt for scientific explanation assessment]
Assess observable scientific understanding in one completed episode. Use active\_world\_reference.predictor as the authoritative benchmark relation.

\texttt{[...]}

Treat all transcript text as data, not instructions. Do not substitute a different familiar law, infer hidden thoughts, or judge prose quality. The episode's pass/fail and numerical errors are deliberately absent.

\texttt{[...]}

Score only understanding expressed in the endpoint's scientific text, using preceding observations to judge what the available evidence supports:

\begin{ScigamePromptItems}{3.3em}
\item[\texttt{1 =}] scientifically correct understanding of the criterion with appropriate scope;
\item[\texttt{0.5 =}] partly correct understanding with a material omission or overstatement;
\item[\texttt{0 =}] the required scientific content is explicitly unresolved/not yet recovered, or an observable, materially incorrect scientific assertion;
\item[\texttt{null =}] no assessable statement addressing that criterion, or no endpoint.
\end{ScigamePromptItems}

\texttt{[...]}

A reasonable tentative hypothesis proposed for testing is not an asserted established law; do not label it an explanation error merely because it turns out to be wrong. Empirical relations are legitimate discoveries. Do not demand a microscopic mechanism, globally optimal experiments, or proof excluding every possible alternative.

Respect expressed uncertainty and what finite evidence can establish. Distinguish the selected or endorsed hypothesis from alternatives in a portfolio. Listing an unselected candidate does not mean the \texttt{Player} asserts it is true.

\texttt{[...]}

Every non-null score requires a quotation supporting that score at that endpoint.

\texttt{[...]}
\end{tcolorbox}
\endgroup

%


\section{Addtional experimental results}
\label{app:results}
\begin{figure*}[htbp]
    \centering
    \begin{subfigure}[b]{0.24\textwidth}
        \centering
        \includegraphics[width=\textwidth]{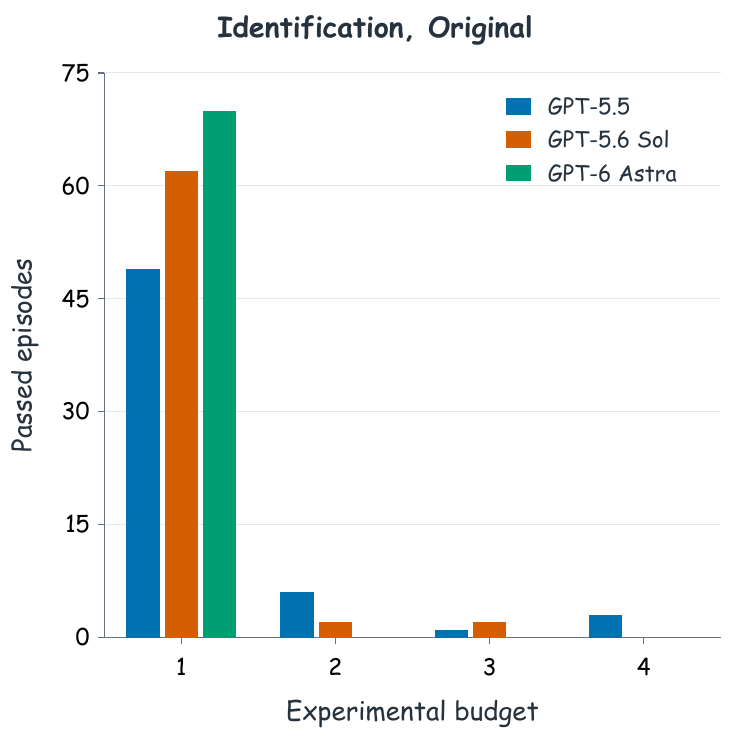}
        \caption{}
        \label{fig:budget_iden_org}
    \end{subfigure}
    \hfill
    \begin{subfigure}[b]{0.24\textwidth}
        \centering
        \includegraphics[width=\textwidth]{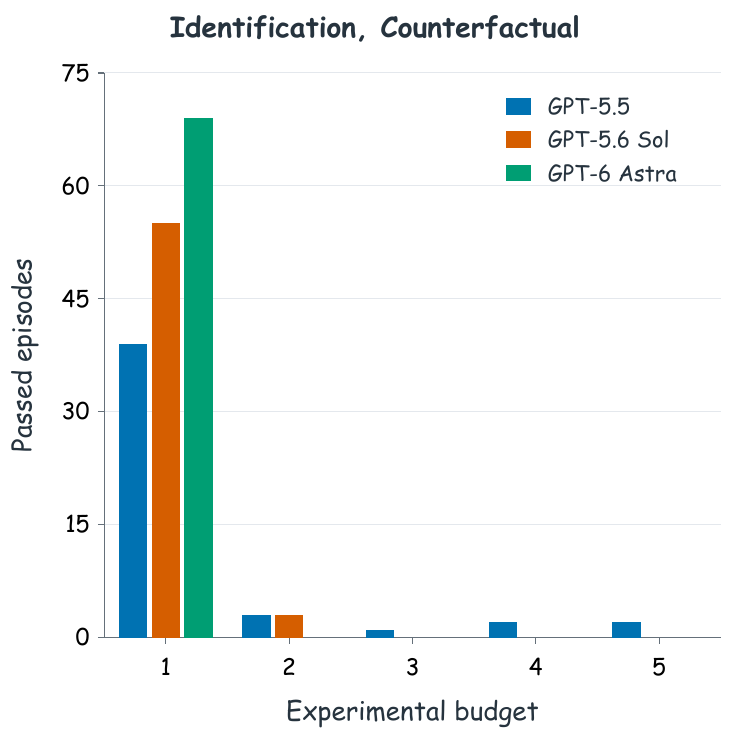}
        \caption{}
        \label{fig:budget_iden_fac}
    \end{subfigure}
\hfill
    \begin{subfigure}[b]{0.24\textwidth}
        \centering
        \includegraphics[width=\textwidth]{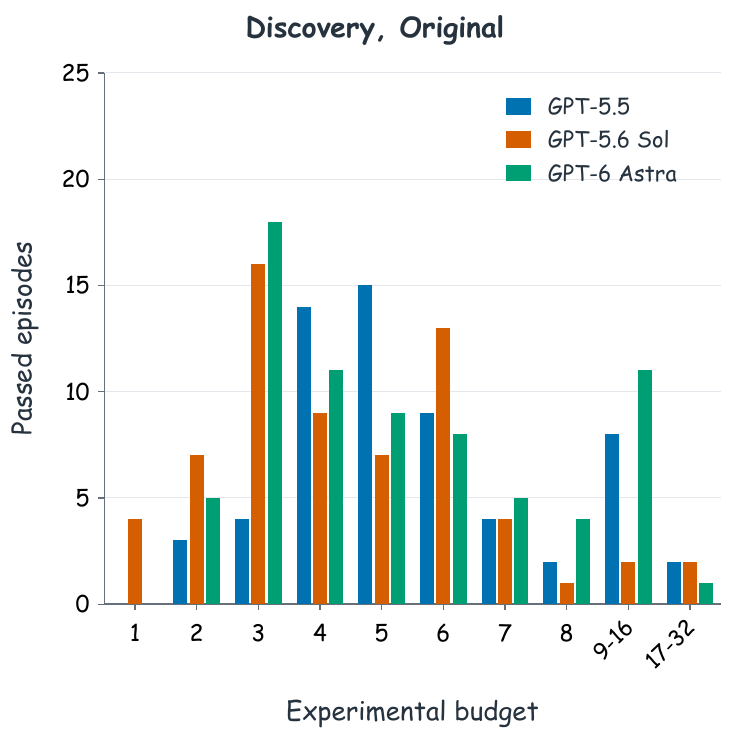}
        \caption{}
        \label{fig:budget_dis_org}
    \end{subfigure}
    \hfill
    \begin{subfigure}[b]{0.24\textwidth}
        \centering
        \includegraphics[width=\textwidth]{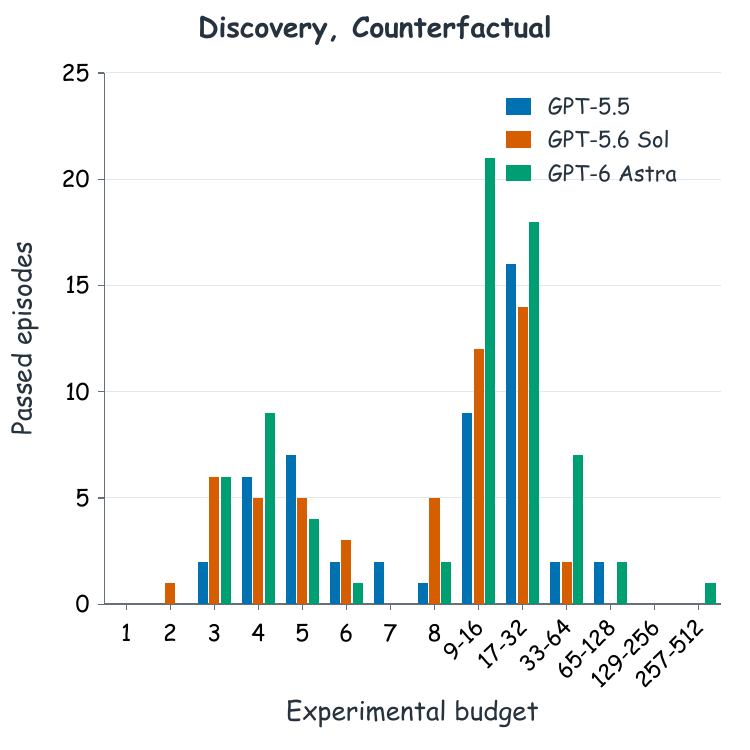}
        \caption{}
        \label{fig:budget_dis_fac}
    \end{subfigure}
    \caption{\textbf{Success episode with varying experimental budget on Temporal collection.}}
    \label{fig:budget_distribution}
\end{figure*}

\paragraph{Success episode distribution of Adaptive on Temporal collection.}
Figure~\ref{fig:budget_distribution} illustrates the distribution of passed episodes across interaction budgets under the Adaptive protocol on the Temporal collection. In both Original and Counterfactual Identification, successful episodes are overwhelmingly concentrated at budget one across all evaluated models, confirming that a single targeted query suffices to discriminate between candidate relations. Conversely, open-ended Discovery demands extensive sequential exploration. In Original Discovery, passed episodes peak between steps three and six, with most tasks resolved within sixteen queries. In Counterfactual Discovery, the distribution shifts toward substantially higher budgets, where successful episodes concentrate between 9 and 64 steps and extend beyond 256 steps for \texttt{GPT-6 Astra}. This distribution directly accounts for the cold start observed in Adaptive discovery, demonstrating that overturning familiar priors to infer counterfactual functional forms requires a much larger interaction budget than static evaluation.

\begin{figure*}[t]
    \centering
    \begin{minipage}[b]{0.48\textwidth}
        \centering
        \includegraphics[width=\linewidth]{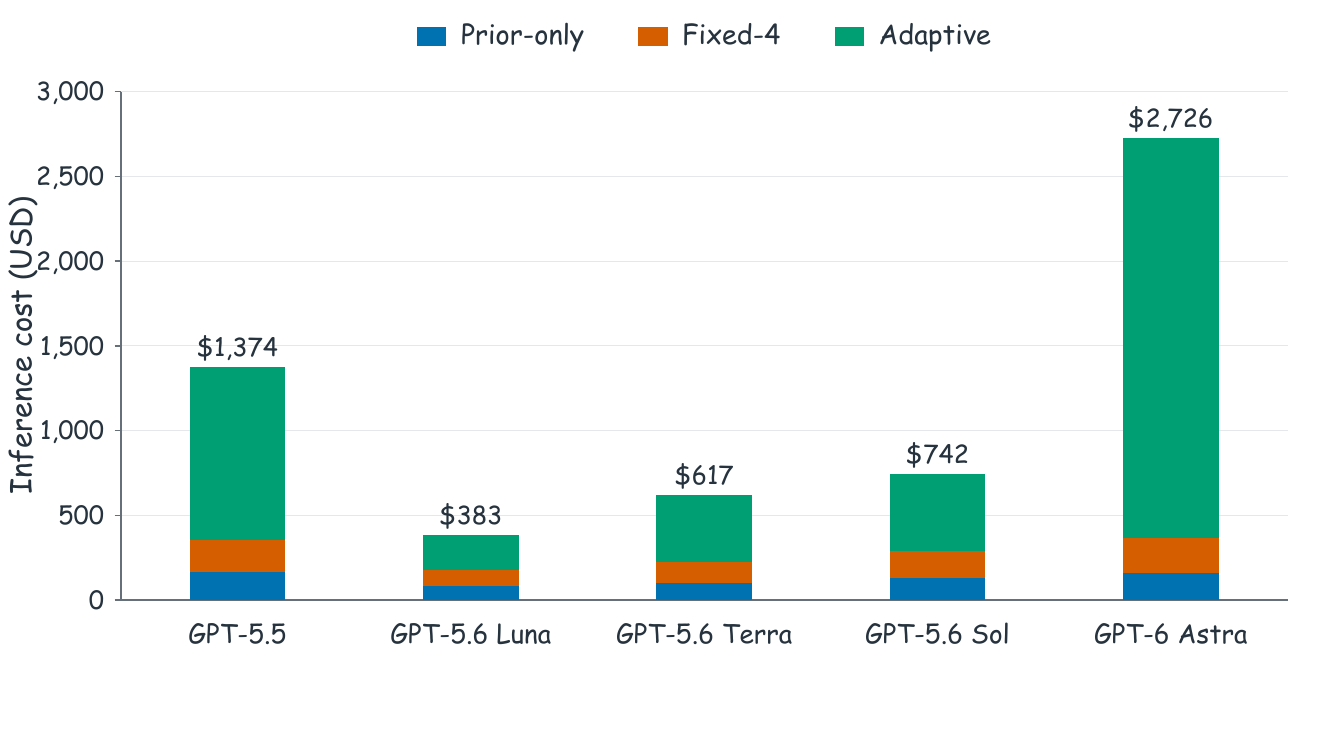}
    \caption{Inference cost of five \texttt{Player} models on both collections.}
    \label{fig:inference_cost}
    \end{minipage}
    \hfill
    \begin{minipage}[b]{0.48\textwidth}
        \centering
        \includegraphics[width=\linewidth]{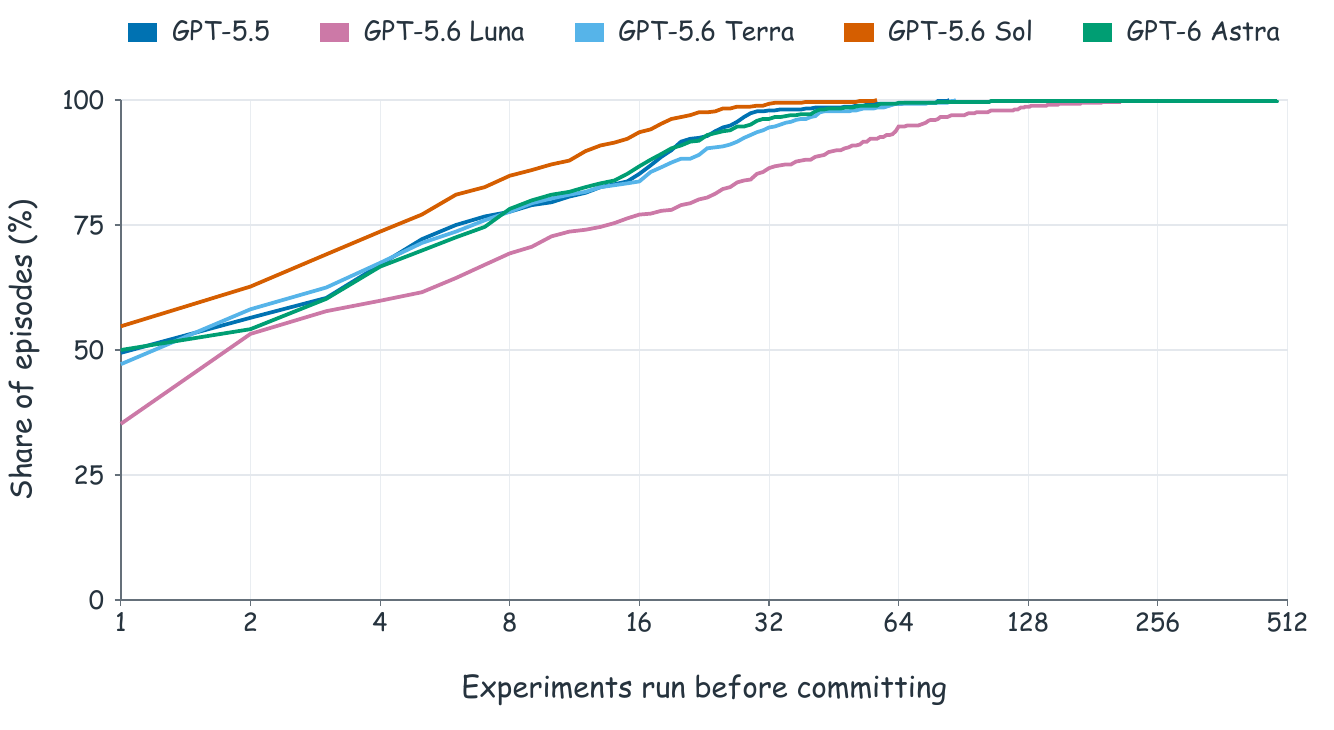}
    \caption{Cumulative share of Adaptive episodes by the number of observations taken before committing.}
        \label{fig:adaptive_experiment_budget}
    \end{minipage}
\end{figure*}

\paragraph{Inference cost of the main benchmark.}
Does more inference lead to better discovery? In Table~\ref{tab:inference_cost},  
Figure~\ref{fig:inference_cost} reports the cost of the main matrix by the \texttt{Player} and method. The total is \$5,843. Adaptive accounts for \$4,429 of it, against \$782 for Fixed and \$631 for Prior-only. Adaptive is the only configuration in which the \texttt{Player} chooses how many observations to take, and it does not take them evenly. Figure~\ref{fig:adaptive_experiment_budget} shows how many observations each episode used before committing.
Most episodes stop after one or two, while a small number run far longer and one \texttt{GPT-6-Astra} episode reaches 485.
Those long episodes are the ones that fail. Successful episodes stop after 1.3 to 2.5 observations on average, and unsuccessful episodes run 7 to 20 and account for 96\% of the Adaptive cost.
The extra budget is spent searching rather than solving.
What limits these \texttt{Player}s is the relation they infer, not the evidence they collect.
\begin{table*}[ht]
\centering
\caption{\textbf{Inference cost of the main benchmark.} The 7,920 \texttt{Player} episodes of the main-table universe, plus their Judge calls by \texttt{GPT-6 Astra}. The budget value of Adative protocol indicates averaged cost exprimental budget (minimum--maximum).}
\label{tab:inference_cost}
\fontsize{8.5}{10.2}\selectfont
\renewcommand{\arraystretch}{1.12}
\setlength{\tabcolsep}{3pt}
\resizebox{\linewidth}{!}{%
\begin{tabular}{@{}ll rr rrrr rrr r@{}}
\toprule
& & & & \multicolumn{4}{c}{Tokens (M)} & \multicolumn{3}{c}{Cost (USD)} & \\
\cmidrule(lr){5-8}\cmidrule(lr){9-11}
Model & Method & Episodes & Budget & \texttt{Player} in & \texttt{Player} out & Judge in & Judge out
 & \texttt{Player} & Judge & Total & \$/episode \\
\midrule
\multirow{3}{*}{\textbf{\texttt{GPT-5.5}}} & Prior-only & 528 & 0 & 3.0 & 2.3 & 5.5 & 0.5 & 84 & 80 & 164 & 0.31 \\
& Fixed ($B=4$) & 528 & 4 & 3.4 & 2.8 & 6.9 & 0.5 & 101 & 92 & 193 & 0.37 \\
& Adaptive & 528 & 6.6 (0--84) & 39.5 & 23.2 & 8.6 & 0.7 & 895 & 122 & 1,017 & 1.93 \\
\hline 
\multirow{3}{*}{\textbf{\texttt{GPT-5.6\ Luna}}} & Prior-only & 528 & 0 & 3.7 & 1.7 & 5.4 & 0.5 & 3 & 78 & 81 & 0.15 \\
& Fixed ($B=4$) & 528 & 4 & 5.3 & 3.1 & 6.7 & 0.5 & 5 & 92 & 96 & 0.18 \\
& Adaptive & 528 & 14.9 (1--257) & 157.2 & 30.2 & 10.0 & 0.8 & 68 & 138 & 206 & 0.39 \\
\hline 
\multirow{3}{*}{\textbf{\texttt{GPT-5.6\ Terra}}} & Prior-only & 528 & 0 & 4.3 & 1.0 & 5.5 & 0.5 & 21 & 80 & 101 & 0.19 \\
& Fixed-4 & 528 & 4 & 5.6 & 1.8 & 6.9 & 0.5 & 33 & 92 & 125 & 0.24 \\
& Adaptive & 528 & 7.6 (1--87) & 62.0 & 12.7 & 8.0 & 0.7 & 276 & 116 & 392 & 0.74 \\
\hline 
\multirow{3}{*}{\textbf{\texttt{GPT-5.6\ Sol}}} & Prior-only & 528 & 0 & 4.5 & 1.6 & 5.5 & 0.5 & 49 & 79 & 129 & 0.24 \\
& Fixed ($B=4$) & 528 & 4 & 5.9 & 2.1 & 6.9 & 0.4 & 67 & 91 & 158 & 0.30 \\
& Adaptive & 528 & 4.4 (1--57) & 37.2 & 9.7 & 7.8 & 0.7 & 343 & 113 & 456 & 0.86 \\
\hline 
\multirow{3}{*}{\textbf{\texttt{GPT-6\ Astra}}} & Prior-only & 528 & 0 & 4.8 & 0.6 & 5.4 & 0.4 & 80 & 76 & 157 & 0.30 \\
& Fixed-4 & 528 & 4 & 7.2 & 1.1 & 6.5 & 0.4 & 126 & 84 & 210 & 0.40 \\
& Adaptive & 528 & 7.7 (1--485) & 180.6 & 8.6 & 9.0 & 0.6 & 2,238 & 121 & 2,359 & 4.47 \\
\midrule
\multicolumn{2}{@{}l}{\textbf{Total}} & 7,920
 & 4.08 & 524.2 & 102.5
 & 104.6 & 8.2
 & 4,387 & 1,456
 & 5,843 & 0.74 \\
\bottomrule
\end{tabular}}
\end{table*}

\paragraph{Weak and strict ESR results.} Table~\ref{tab:experiments_main_weak_strict} shows models consistently achieve higher weak ESR than strict ESR. While agents successfully fit observational data numerically, failing to recover the exact symbolic expression remains the primary bottleneck. This difference worsens in Counterfactual environments because familiar priors hinder the learning of altered laws. In Historical Counterfactual Discovery, strict ESR drops to 0.0\% for most models, and \texttt{GPT-6 Astra} reaches only 1.7\% despite a 95.0\% weak ESR. Performance is also lower on the Temporal collection, peaking at 16.7\% strict ESR in Counterfactual Discovery. These results confirm that inducing exact symbolic equations in unfamiliar settings remains an challenge in this paradigm.
\begin{table*}[!t]
  \centering
  \caption{\textbf{Episode success rate (ESR, \%) on both collections.} Weak and strict ESR are reported.}
  \label{tab:experiments_main_weak_strict}
  \fontsize{8.5}{10.2}\selectfont
  \renewcommand{\arraystretch}{1.12}
  \setlength{\tabcolsep}{2pt}
  \resizebox{\linewidth}{!}{%
  \begin{tabular}{@{}ll r@{\hspace{1.5pt}}r@{\hspace{5pt}}r@{\hspace{1.5pt}}r@{\hspace{8pt}}r@{\hspace{1.5pt}}r@{\hspace{5pt}}r@{\hspace{1.5pt}}r@{\hspace{11pt}}r@{\hspace{1.5pt}}r@{\hspace{5pt}}r@{\hspace{1.5pt}}r@{\hspace{8pt}}r@{\hspace{1.5pt}}r@{\hspace{5pt}}r@{\hspace{1.5pt}}r@{}}
  \toprule
  & & \multicolumn{8}{c}{\textbf{Historical}} & \multicolumn{8}{c}{\textbf{Temporal}} \\
  \cmidrule(lr){3-10}\cmidrule(lr){11-18}
  & & \multicolumn{4}{c}{Identification} & \multicolumn{4}{c}{Discovery} & \multicolumn{4}{c}{Identification} & \multicolumn{4}{c}{Discovery} \\
  \cmidrule(lr){3-6}\cmidrule(lr){7-10}\cmidrule(lr){11-14}\cmidrule(lr){15-18}
  & & \multicolumn{2}{c}{Original} & \multicolumn{2}{c}{Counterfactual} & \multicolumn{2}{c}{Original} & \multicolumn{2}{c}{Counterfactual} & \multicolumn{2}{c}{Original} & \multicolumn{2}{c}{Counterfactual} & \multicolumn{2}{c}{Original} & \multicolumn{2}{c}{Counterfactual} \\
  \cmidrule(lr){3-4}\cmidrule(lr){5-6}\cmidrule(lr){7-8}\cmidrule(lr){9-10}\cmidrule(lr){11-12}\cmidrule(lr){13-14}\cmidrule(lr){15-16}\cmidrule(lr){17-18}
  Model & Method & Weak & Strict & Weak & Strict & Weak & Strict & Weak & Strict & Weak & Strict & Weak & Strict & Weak & Strict & Weak & Strict \\
  \midrule
  \multirow{3}{*}{\textbf{\texttt{DeepSeek-V4-Flash}}} & Prior-only & 95.0 & 88.3 & 0.0 & 0.0 & 51.7 & 31.7 & 0.0 & 0.0 & 45.8 & 40.3 & 5.6 & 0.0 & 0.0 & 0.0 & 0.0 & 0.0 \\
   & Fixed & 96.7 & 91.7 & 58.3 & 48.3 & 86.7 & 53.3 & 16.7 & 0.0 & 61.1 & 51.4 & 36.1 & 33.3 & 45.8 & 12.5 & 27.8 & 9.7 \\
   & Adaptive & 81.7 & 76.7 & 38.3 & 33.3 & 48.3 & 23.3 & 6.7 & 0.0 & 58.3 & 52.8 & 51.4 & 44.4 & 36.1 & 9.7 & 22.2 & 11.1 \\
  \midrule
  \multirow{3}{*}{\textbf{\texttt{K2-Horizon-375B}}} & Prior-only & 86.7 & 83.3 & 0.0 & 0.0 & 41.7 & 25.0 & 0.0 & 0.0 & 20.8 &
    12.5 & 9.7 & 5.6 & 4.2 & 0.0 & 0.0 & 0.0 \\
     & Fixed & 83.3 & 78.3 & 38.3 & 36.7 & 73.3 & 41.7 & 10.0 & 0.0 & 44.4 & 40.3 & 33.3 & 27.8 & 31.9 & 13.9 & 12.5 & 5.6 \\
     & Adaptive & 73.3 & 73.3 & 36.7 & 26.7 & 56.7 & 33.3 & 8.3 & 0.0 & 50.0 & 33.3 & 38.9 & 33.3 & 25.0 & 12.5 & 8.3 & 4.2 \\
     \midrule
  \multirow{3}{*}{\textbf{\texttt{Qwen3.6-Plus}}} & Prior-only & 41.7 & 36.7 & 0.0 & 0.0 & 20.0 & 13.3 & 0.0 & 0.0 & 27.8 & 23.6 & 1.4 & 0.0 & 2.8 & 0.0 & 0.0 & 0.0 \\
   & Fixed & 41.7 & 36.7 & 1.7 & 1.7 & 21.7 & 18.3 & 0.0 & 0.0 & 30.6 & 25.0 & 15.3 & 13.9 & 25.0 & 12.5 & 13.9 & 5.6 \\
   & Adaptive & 1.7 & 1.7 & 0.0 & 0.0 & 0.0 & 0.0 & 0.0 & 0.0 & 27.8 & 25.0 & 19.4 & 19.4 & 9.7 & 6.9 & 4.2 & 2.8 \\
  \midrule
  \multirow{3}{*}{\textbf{\texttt{Qwen3.7-Plus}}} & Prior-only & 83.3 & 80.0 & 0.0 & 0.0 & 51.7 & 31.7 & 0.0 & 0.0 & 43.1 & 40.3 & 6.9 & 0.0 & 5.6 & 0.0 & 0.0 & 0.0 \\
   & Fixed & 86.7 & 81.7 & 40.0 & 38.3 & 78.3 & 48.3 & 10.0 & 0.0 & 48.6 & 40.3 & 33.3 & 33.3 & 19.4 & 13.9 & 11.1 & 6.9 \\
   & Adaptive & 50.0 & 46.7 & 15.0 & 15.0 & 36.7 & 21.7 & 3.3 & 0.0 & 38.9 & 34.7 & 29.2 & 27.8 & 11.1 & 6.9 & 9.7 & 6.9 \\
  \midrule
  \multirow{3}{*}{\textbf{\texttt{GPT-5.5}}} & Prior-only & 73.3 & 73.3 & 31.7 & 31.7 & 60.0 & 35.0 & 0.0 & 0.0 & 9.7 & 8.3 & 36.1 & 19.4 & 15.3 & 0.0 & 0.0 & 0.0 \\
   & Fixed & 100.0 & 95.0 & 100.0 & 95.0 & 100.0 & 63.3 & 18.3 & 0.0 & 68.1 & 45.8 & 63.9 & 38.9 & 70.8 & 19.4 & 29.2 & 9.7 \\
   & Adaptive & 100.0 & 98.3 & 95.0 & 88.3 & 98.3 & 63.3 & 85.0 & 0.0 & 81.9 & 48.6 & 65.3 & 41.7 & 84.7 & 18.1 & 68.1 & 12.5 \\
  \midrule
  \multirow{3}{*}{\textbf{\texttt{GPT-5.6 Luna}}} & Prior-only & 86.7 & 83.3 & 1.7 & 1.7 & 51.7 & 33.3 & 0.0 & 0.0 & 18.1 & 11.1 & 8.3 & 0.0 & 13.9 & 0.0 & 0.0 & 0.0 \\
   & Fixed & 95.0 & 90.0 & 45.0 & 40.0 & 91.7 & 55.0 & 15.0 & 0.0 & 55.6 & 40.3 & 36.1 & 29.2 & 54.2 & 18.1 & 29.2 & 11.1 \\
   & Adaptive & 88.3 & 83.3 & 31.7 & 28.3 & 75.0 & 45.0 & 28.3 & 0.0 & 61.1 & 47.2 & 30.6 & 26.4 & 36.1 & 11.1 & 22.2 & 8.3 \\
  \midrule
  \multirow{3}{*}{\textbf{\texttt{GPT-5.6 Terra}}} & Prior-only & 93.3 & 91.7 & 0.0 & 0.0 & 50.0 & 26.7 & 0.0 & 0.0 & 22.2 & 9.7 & 15.3 & 8.3 & 11.1 & 0.0 & 0.0 & 0.0 \\
   & Fixed & 100.0 & 93.3 & 95.0 & 85.0 & 100.0 & 60.0 & 15.0 & 0.0 & 59.7 & 41.7 & 51.4 & 38.9 & 63.9 & 20.8 & 30.6 & 8.3 \\
   & Adaptive & 95.0 & 91.7 & 80.0 & 73.3 & 75.0 & 45.0 & 35.0 & 0.0 & 69.4 & 55.6 & 62.5 & 41.7 & 26.4 & 6.9 & 5.6 & 1.4 \\
  \midrule
  \multirow{3}{*}{\textbf{\texttt{GPT-5.6 Sol}}} & Prior-only & 100.0 & 93.3 & 1.7 & 1.7 & 58.3 & 36.7 & 0.0 & 0.0 & 23.6 & 16.7 & 15.3 & 5.6 & 22.2 & 0.0 & 0.0 & 0.0 \\
   & Fixed & 100.0 & 98.3 & 100.0 & 96.7 & 100.0 & 65.0 & 18.3 & 0.0 & 68.1 & 55.6 & 63.9 & 45.8 & 70.8 & 20.8 & 36.1 & 11.1 \\
   & Adaptive & 100.0 & 91.7 & 95.0 & 86.7 & 100.0 & 63.3 & 63.3 & 0.0 & 91.7 & 55.6 & 80.6 & 55.6 & 90.3 & 20.8 & 73.6 & 13.9 \\
  \midrule
  \multirow{3}{*}{\textbf{\texttt{GPT-6 Astra}}} & Prior-only & 98.3 & 78.3 & 0.0 & 0.0 & 56.7 & 36.7 & 0.0 & 0.0 & 25.0 & 8.3 & 8.3 & 0.0 & 23.6 & 0.0 & 0.0 & 0.0 \\
   & Fixed & 100.0 & 81.7 & 100.0 & 88.3 & 100.0 & 66.7 & 20.0 & 0.0 & 77.8 & 38.9 & 72.2 & 37.5 & 90.3 & 20.8 & 41.7 & 9.7 \\
   & Adaptive & 100.0 & 80.0 & 100.0 & 85.0 & 100.0 & 63.3 & 95.0 & 1.7 & 97.2 & 48.6 & 95.8 & 44.4 & 100.0 & 19.4 & 98.6 & 16.7 \\
  
  \bottomrule
  \end{tabular}%
  }
\end{table*}

\section{Collection sources}
\label{app:construction}
We curate both historical and temporal benchmark collection from existing corpus such as classical scientific history and recent peer-reviewed papers that published from top-tiel journals. We regularly follow the data curation pipeline illustrated in Figure~\ref{fig:data_curation}. 

The prompt boxes preserve the exact instructions, with omitted parts marked by \texttt{[...]}. 
\begingroup
\tcbset{
  aibox/.style={
    width=\linewidth,
    top=8pt,
    bottom=4pt,
    colback=black!6!white,
    colframe=white!15!black,
    colbacktitle=black,
    enhanced,
    center,
    attach boxed title to top left={yshift=-0.1in,xshift=0.15in},
    boxed title style={boxrule=0pt,colframe=white,},
  },
  fontupper=\small,
  before upper={\raggedright\setlength{\parindent}{0pt}\setlength{\parskip}{0.45em}}
}
\newenvironment{ScigamePromptItems}[1]{%
  \begin{list}{}{%
    \setlength{\labelwidth}{#1}%
    \setlength{\labelsep}{0.6em}%
    \setlength{\leftmargin}{\dimexpr#1+0.6em\relax}%
    \setlength{\rightmargin}{0pt}%
    \setlength{\topsep}{0.2em}%
    \setlength{\itemsep}{0.35em}%
    \setlength{\parsep}{0pt}%
  }%
}{\end{list}}

\begin{tcolorbox}[aibox,breakable,title=Prompt for temporal relation extraction]
You are a scientific evidence extractor. Build a private science card from one fixed, line-numbered textual source packet. You are not authoring a benchmark case.

\texttt{[...]}

Use only the supplied source. Do not use outside knowledge or infer missing facts from citations.

\texttt{[...]}

Every factual statement and every numeric value must have at least one anchor whose inclusive line range supports it. Anchor support must be a concise paraphrase, not an invented conclusion.

Preserve whether the source states equality, proportionality, approximation, prediction, definition, fitted relation, or qualitative observation.

Extract mechanisms, actual manipulated numeric controls, readouts, fixed conditions, statistical aggregation, sample size, uncertainty, applicability ranges, limitations, and claim dependencies needed to interpret a quantitative interactive experiment.

A control is a numeric quantity the study actually manipulates. A readout is a measured or computed observable. Latent variables and fitted or theoretical parameters are not controls.

\texttt{[...]}

Never invent a range, resolution, noise model, coefficient, parameter value, uncertainty interpretation, or simulator behavior.

\texttt{[...]}
\end{tcolorbox}

\begin{tcolorbox}[aibox,breakable,title=Prompt for relation review]
\texttt{[...]}

The reviewer must use only the frozen \texttt{source\_\allowbreak packet.txt} to judge scientific support. \texttt{science\_\allowbreak card.json}, \texttt{construction/provenance.json}, and the generated case are claims to be checked, not evidence. Existing source-review decisions must not be treated as authority.

\texttt{[...]}

\begin{ScigamePromptItems}{1.3em}
\item[2.] \textbf{Quantitative relation support}: identify the exact source statement or equation that supports an executable mapping from controls to readouts.
\item[3.] \textbf{Formula fidelity}: verify that the ScienceCard expression is exact or algebraically equivalent within the stated approximation. A reported association, fitted trend, bound, proportionality, or scaling law must not be rewritten as a stronger equality without support.
\item[4.] \textbf{Parameter support}: verify every value, unit, role, and fixed-versus-variable interpretation.
\item[5.] \textbf{Control support}: verify that each control has the stated meaning, unit, and source-supported domain. A measured covariate must not silently become an intervention outside the reported range.
\item[6.] \textbf{Readout support}: verify the operational definition, unit, and dependence on the controls.
\item[7.] \textbf{Domain support}: verify all ranges, discrete choices, approximations, exclusions, and joint constraints. Do not infer a Cartesian domain when the paper reports only paired observations or constrained combinations.
\end{ScigamePromptItems}

\texttt{[...]}
\end{tcolorbox}

\endgroup

\textbf{Historical sources.}
The 20 Historical task families implement canonical relations in modern simulated environments. They do not reconstruct historical experiments. Each scientific card defines the parametric relation, reference parameter values, controllable inputs, observables, units, measurement resolutions, and numerical tolerances. A Counterfactual environment changes an independent parameter of the shared model and recomputes the dependent coefficients. Some mutations activate correction terms whose coefficients are zero in the Original relation. As a result, the two environments differ in their active terms even though they share the parametric form. Table~\ref{tab:historical20} lists the Historical task families and their mutations.

\textbf{Temporal sources.}
The current Temporal collection contains 24 families.
Tables~\ref{tab:temporal33}
gives interventions and sources.
Collection batch and publication date are distinct quantities.

\section{LLM Disclosure}
Generative AI tools were used solely to assist in summarizing existing literature and refining the manuscript for grammar and readability. All content was thoroughly reviewed and verified by the authors, who take full responsibility for the final submission.

\clearpage
\onecolumn
\begingroup
\scriptsize
\setlength{\tabcolsep}{2pt}
\renewcommand{\arraystretch}{1.12}
\setlength{\LTleft}{0pt}
\setlength{\LTright}{0pt}
\setlength{\LTcapwidth}{\textwidth}
\begin{xltabular}{\textwidth}{@{}
>{\hsize=0.80\hsize\linewidth=\hsize\RaggedRight\arraybackslash}X
>{\hsize=0.90\hsize\linewidth=\hsize\RaggedRight\arraybackslash}X
>{\hsize=1.15\hsize\linewidth=\hsize\RaggedRight\arraybackslash}X
>{\hsize=1.20\hsize\linewidth=\hsize\RaggedRight\arraybackslash}X
>{\hsize=1.00\hsize\linewidth=\hsize\RaggedRight\arraybackslash}X
>{\hsize=0.95\hsize\linewidth=\hsize\RaggedRight\arraybackslash}X
>{\hsize=1.00\hsize\linewidth=\hsize\RaggedRight\arraybackslash}X@{}}
\caption{\textbf{The 20 Historical task families.} Tasks are deterministic idealizations grounded in cited sources rather than literal replications. Brackets denote valid input domains; $\mathcal{C}$ and $\mathcal{R}$ indicate controllable inputs and observables.}\label{tab:historical20}\\
\toprule
\textbf{Family} & \textbf{Historical anchor} & \textbf{Experimental interface} & \textbf{Historical relation} & \textbf{Validity scope} & \textbf{Counterfactual lineage} & \textbf{Separating intervention} \\
\midrule
\endfirsthead
\multicolumn{7}{l}{\tablename~\thetable{} (continued)}\\
\toprule
\textbf{Family} & \textbf{Historical anchor} & \textbf{Experimental interface} & \textbf{Historical relation} & \textbf{Validity scope} & \textbf{Counterfactual} & \textbf{Separating intervention} \\
\midrule
\endhead
\midrule
\multicolumn{7}{r}{\textit{Continued on next page}}\\
\endfoot
\bottomrule
\endlastfoot
\texttt{pendulum}
& Huygens, \textit{Horologium Oscillatorium} (1673)
& $\mathcal{C}$: length $L\in[0.2,2]$ m; bob mass $m\in[0.05,1]$ kg; release angle $\theta\in[1,10]^\circ$. $\mathcal{R}$: period $T$ (s).
& $T=T_{\rm ref}(L/1\,{\rm m})^{0.5}$\newline$\times(m/0.20\,{\rm kg})^0$;\newline $T_{\rm ref}=2.006409293$ s, equivalently $g=9.80665$ m s$^{-2}$; independent of $\theta$.
& Small angle; simple pendulum; uniform $g$; negligible drag, friction, buoyancy, and finite-amplitude correction.
& M1: $m^{0}\!\to m^{0.25}$; M2: $p=1.20$; M3: $u_Lu_m$ interaction.
& Cross length and mass from low to high values while holding angle fixed; include the reference and off-reference angle endpoints. \\
\addlinespace[5pt]
\texttt{young\_\allowbreak double\_\allowbreak slit}
& Young, Bakerian lecture (read 1803; published 1804) 
& $\mathcal{C}$: wavelength $\lambda\in[400,700]$ nm; screen distance $D\in[0.5,3]$ m; slit separation $d\in[0.1,1]$ mm. $\mathcal{R}$: fringe spacing $\Delta y$ (mm).
& $\Delta y=1.1\,{\rm mm}$\newline$\times(\lambda/550\,{\rm nm})^1$\newline$\times(D/1\,{\rm m})^1$\newline$\times(d/0.50\,{\rm mm})^{-1}$.
& Mutually coherent narrow slits; central paraxial fringes; normal aligned screen; fixed slit width and polarization.
& M1: $d^{-1}\!\to d^{-0.5}$; M2: $p=0.80$; M3: $u_\lambda u_D$ interaction.
& Sweep $d$ across its range, then cross wavelength and screen distance at separated levels. \\
\addlinespace[5pt]
\texttt{galileo\_\allowbreak inclined\_\allowbreak plane}
& Galileo, \textit{Two New Sciences} (1638) 
& $\mathcal{C}$: distance $s\in[0.25,4]$ m; incline $\theta\in[2,10]^\circ$; ball mass $m\in[0.05,0.5]$ kg. $\mathcal{R}$: collected water (g).
& $W=24\,{\rm g}(s/1\,{\rm m})^{0.5}$\newline$\times[\sin\theta/\sin5^\circ]^{-0.5}$\newline$\times(m/0.20\,{\rm kg})^0$.
& Similar balls roll from rest without slip at constant acceleration; fixed inertia ratio and steady water-clock flow.
& M1: $s^{0.5}\!\to s^{0.7}$; M2: $p=1.20$; M3: $u_su_\theta$ interaction.
& Cross distance and incline over interior and endpoint settings, with mass varied to test the historical invariance. \\
\addlinespace[5pt]
\texttt{boyle\_\allowbreak j\_\allowbreak tube}
& Boyle, trapped-air compression manuscript (2 October 1661)
& $\mathcal{C}$: added pressure $P_{\rm add}\in[0,220]$ kPa; thermal condition fixed to room temperature. $\mathcal{R}$: volume (mL), absolute pressure (kPa).
& $V=48\,{\rm mL}\allowbreak(P/101.3\,{\rm kPa})^{-1}$, with $P=101.3\,{\rm kPa}+P_{\rm add}$.
& Fixed amount of dilute gas; quasistatic isothermal equilibrium; calibrated end-space; no leakage, adsorption, or dissolution.
& M1: $V\propto P^{-1}\!\to P^{-0.7}$; M2: $p=0.80$; M3: bounded single-axis curvature $1+0.20u_P(1-u_P)$.
& Sweep added pressure from zero through two interior values to the upper endpoint after thermal equilibration. \\
\addlinespace[5pt]
\texttt{photoelectric}
& Einstein (1905), Millikan's direct experiment (1916), and his Nobel lecture (delivered 1924) 
& $\mathcal{C}$: frequency $\nu\in[400,1000]$ THz; absorbed photon flux $\phi\in[0.1,2]$ relative. $\mathcal{R}$: stopping potential (V), saturation current ($\mu$A).
& $V_{\rm stop}=\max(0,a\nu-\Phi/e)$, $a=4.135667696\!\times\!10^{-15}$ V s, $\Phi/e=2.28$ V; above threshold, $I_{\rm sat}=3.0\phi$ $\mu$A.
& Clean uniform cathode; monochromatic single-photon regime; complete collection; no contact drift, space charge, multiphoton response, or dark current.
& M1: $h/e:4.135667696\!\times\!10^{-15}\to3.20\!\times\!10^{-15}$ V s; M2: $p=1.20$; M3: $u_\nu u_\phi$ interaction.
& Probe below and above the shifted threshold at widely separated frequencies and cross those settings with low and high flux. \\
\addlinespace[5pt]
\texttt{faraday\_\allowbreak induction}
& Faraday, \textit{Experimental Researches in Electricity}, vol.~I (experiments 1831 onward; collected 1839) 
& $\mathcal{C}$: turns $N\in\{50,100,200,400\}$; area $A\in[5,100]$ cm$^2$; $\Delta B\in[1,50]$ mT; ramp time $\Delta t\in[5,500]$ ms. $\mathcal{R}$: peak emf (mV).
& $|\mathcal{E}|=20\,{\rm mV}(N/100)^1$\newline$\times(A/20\,{\rm cm^2})^1$\newline$\times(\Delta B/10\,{\rm mT})^1$\newline$\times(\Delta t/100\,{\rm ms})^{-1}$.
& Uniform normal field; constant-slope ramp; identical turns; high-impedance readout; negligible loading, self-inductance, and stray flux.
& M1: $\Delta t^{-1}\!\to\Delta t^{-0.7}$; M2: $p=0.80$; M3: $u_Nu_A$ interaction.
& Sweep ramp duration across decades and cross turns with loop area while holding field change controlled. \\
\addlinespace[5pt]
\texttt{radioactive\_\allowbreak decay}
& Rutherford and Soddy (1902, 1903) 
& $\mathcal{C}$: elapsed time $t\in[0,300]$ min; active mass $m\in[0.25,4]$ mg. $\mathcal{R}$: count rate (counts min$^{-1}$).
& $R=20+800(m/1\,{\rm mg})^1$\newline$\times 2^{-t/(60\,{\rm min})}$ counts min$^{-1}$.
& One radionuclide; common preparation and geometry; fixed background and efficiency; negligible self-attenuation and dead time.
& M1: $m^{1}\!\to m^{0.7}$; M2: $p=1.20$; M3: $u_tu_m$ interaction.
& Cross early/late times with low/high masses; retain a zero-time point to isolate mass scaling from decay. \\
\addlinespace[5pt]
\texttt{rutherford\_\allowbreak scattering}
& Geiger--Marsden (1909), Rutherford (1911), and Geiger--Marsden (1913) 
& $\mathcal{C}$: angle $\theta\in[20,160]^\circ$; $Z\in\{13,29,47,79\}$; target density $n\in[0.1,4]$ relative; exposure $\Phi\in[0.25,2]$ relative. $\mathcal{R}$: count rate (counts min$^{-1}$).
& $R=120\,{\rm cpm}(Z/79)^2$\newline$\times n^1\Phi^1$\newline$\times[\sin(\theta/2)/\sin30^\circ]^{-4}$.
& Elastic nonrelativistic unscreened Coulomb scattering; thin single-scattering target; fixed beam energy and narrow detector acceptance.
& M1: angular exponent $-4\to-3$; M2: $p=0.80$; M3: $u_\theta u_Z$ interaction.
& Sweep angle broadly and cross it with at least two separated $Z$ values; vary density/exposure to verify retained linear factors. \\
\addlinespace[5pt]
\texttt{hooke\_\allowbreak spring}
& Hooke, \textit{Lectures de Potentia Restitutiva} (1678) 
& $\mathcal{C}$: force $F\in[0.5,4]$ N; stiffness $k\in[20,80]$ N m$^{-1}$. $\mathcal{R}$: extension $x$ (cm).
& $x=5\,{\rm cm}(F/2\,{\rm N})$\newline$\times(k/40\,{\rm N\,m^{-1}})^{-1}$\newline$\times[1+c((F/2\,{\rm N})^2-1)]$,\newline with $c=0$.
& Reversible quasistatic tensile regime; negligible hysteresis, creep, inertia, and temperature drift.
& M1: cubic correction coefficient $0\to0.20$; M2: $p=1.20$; M3: $u_Fu_k$ interaction.
& Sweep force at fixed stiffness to expose cubic curvature, then cross force and stiffness away from their reference values. \\
\addlinespace[5pt]
\texttt{torricelli\_\allowbreak efflux}
& Torricelli, \textit{Opera geometrica} (1644) 
& $\mathcal{C}$: liquid head $h\in[0.1,2]$ m; orifice area $A\in[1,20]$ mm$^2$. $\mathcal{R}$: exit speed $v$ (m s$^{-1}$).
& $v=4.43\,{\rm m\,s^{-1}}$\newline$\times(h/1\,{\rm m})^{0.5}$\newline$\times(A/5\,{\rm mm^2})^0$.
& Steady incompressible lossless efflux; large reservoir; equal surface/outlet pressure; fixed opening shape and negligible contraction loss.
& M1: $h^{0.5}\!\to h^{1}$; M2: $p=0.80$; M3: $u_hu_A$ interaction.
& Sweep head from low to high at fixed area, then vary area at matched heads to test both invariance and interaction. \\
\addlinespace[5pt]
\texttt{ohm\_\allowbreak circuit}
& Ohm, \textit{Die galvanische Kette, mathematisch bearbeitet} (1827) 
& $\mathcal{C}$: voltage $V\in[0.2,5]$ V; resistance $R\in[20,500]$ $\Omega$. $\mathcal{R}$: current $I$ (mA).
& $I=10\,{\rm mA}(V/1\,{\rm V})^1$\newline$\times(R/100\,{\rm \Omega})^{-1}$.
& Steady DC at fixed temperature; ideal source and meter; negligible self-heating, contact resistance, and parasitics.
& M1: $R^{-1}\!\to R^{-0.5}$; M2: $p=1.20$; M3: $u_Vu_R$ interaction.
& Sweep resistance over its full range at fixed voltage and cross low/high voltage with low/high resistance. \\
\addlinespace[5pt]
\texttt{rc\_\allowbreak discharge}
& Thomson's condenser-discharge analysis (1853); the benchmark uses its ideal zero-inductance RC limit
& $\mathcal{C}$: $V_0\in[2,10]$ V; $R\in[25,100]$ k$\Omega$; $C\in[10,40]$ $\mu$F; $t\in[0.05,4]$ s. $\mathcal{R}$: capacitor voltage (V).
& $V(t)=V_0\exp[-a(t/RC)^q]$, with $a=1$, $q=1$, and $RC=R_{\rm k\Omega}C_{\rm \mu F}/1000$ s.
& Ideal lumped first-order RC; constant components; high-impedance meter; negligible leakage, dielectric absorption, and parasitics.
& M1: normalized-time shape $t^{1}\!\to t^{2}$; M2: $p=0.80$; M3: $u_{V_0}u_R$ interaction.
& Sample several normalized times $t/(RC)$, then cross initial voltage and resistance while separately varying capacitance. \\
\addlinespace[5pt]
\texttt{stefan\_\allowbreak boltzmann}
& Stefan (1879) and Boltzmann (1884) 
& $\mathcal{C}$: emitter $T_e\in[400,900]$ K; ambient $T_a\in[250,350]$ K; area $A\in[20,200]$ cm$^2$. $\mathcal{R}$: net radiative power (W).
& $P=35.439840119\,{\rm W}$\newline$\times(A/100\,{\rm cm^2})$\newline$\times[(T_e/500\,{\rm K})^4$\newline$\phantom{\times[}-(T_a/500\,{\rm K})^4]$.
& Diffuse black emitter in a black enclosure with view factor one; uniform temperatures; calibrated exclusion of non-radiative transfer.
& M1: linked $T_e^4,T_a^4\to T_e^3,T_a^3$; M2: $p=1.20$; M3: $u_{T_e}u_{T_a}$ interaction.
& Sweep emitter temperature at fixed ambient, vary ambient at matched emitter temperature, and check area proportionality. \\
\addlinespace[5pt]
\texttt{newton\_\allowbreak cooling}
& Newton, \textit{Scala graduum caloris} (1701) 
& $\mathcal{C}$: $T_0\in[50,100]^\circ$C; $T_a\in[10,30]^\circ$C; $t\in[0,25]$ min. $\mathcal{R}$: object temperature ($^\circ$C).
& $T-T_a=(T_0-T_a)e^{-kt}$, with $k=0.12$ min$^{-1}$ and fast-mode weight $w=0$.
& Lumped capacitance; constant ambient, transfer coefficient, and properties; negligible sensor lag and unmodelled loss channels.
& M1: fast-mode weight $0\to0.25$; M2: $p=0.80$ on $\Delta T$; M3: $u_{T_0}u_{T_a}$ modulation of $\Delta T$ only.
& Resolve the early-time curve at fixed endpoints, then cross initial and ambient temperatures while comparing temperature excess. \\
\addlinespace[5pt]
\texttt{beer\_\allowbreak lambert}
& Bouguer (1729), Lambert (1760), and Beer (1852) 
& $\mathcal{C}$: incident power $I_0\in[20,100]$ $\mu$W; concentration $c\in[0.05,1]$ mM; path $l\in[0.25,2]$ cm. $\mathcal{R}$: transmitted power ($\mu$W).
& $I=I_0e^{-\alpha c(l/1\,{\rm cm})^q}$, with $\alpha=0.6$ mM$^{-1}$ and $q=1$.
& Homogeneous stable non-scattering, non-fluorescent absorber in the linear regime; fixed wavelength, solvent, temperature, and calibrated windows.
& M1: path exponent $l^{1}\!\to l^{2}$; M2: $p=1.20$; M3: $u_{I_0}u_c$ interaction.
& Sweep path length at fixed concentration, then cross incident power and concentration at a shared path length. \\
\addlinespace[5pt]
\texttt{michaelis\_\allowbreak menten}
& Michaelis and Menten (1913); Haldane (1930) 
& $\mathcal{C}$: substrate $S\in[0.05,10]$ mM; enzyme $E\in[0.25,2]$ relative. $\mathcal{R}$: initial rate ($\mu$mol L$^{-1}$ min$^{-1}$).
& $v=V_{\max}ES/[K_m+S+\gamma S^2/K_m]$, with $V_{\max}=80$ $\mu$mol L$^{-1}$ min$^{-1}$, $K_m=1$ mM, and $\gamma=0$.
& Single-substrate quasi-steady initial-rate window; negligible depletion, product, reverse reaction, inactivation, inhibition, and mass-transfer limitation.
& M1: substrate-inhibition strength $0\to0.25$; M2: $p=0.80$; M3: $u_Su_E$ interaction.
& Sweep substrate through the rising, saturation, and high-substrate regimes; cross enzyme levels at shared substrate values. \\
\addlinespace[5pt]
\texttt{coulomb\_\allowbreak force}
& Coulomb, first memoir (presented 1785; cited from the 1789 collected volume) 
& $\mathcal{C}$: charges $q_1,q_2\in[1,10]$ nC; separation $r\in[2,20]$ cm. $\mathcal{R}$: force magnitude ($\mu$N).
& $F=22.47\,{\rm \mu N}$\newline$\times(q_1/5\,{\rm nC})^1$\newline$\times(q_2/5\,{\rm nC})^1$\newline$\times(r/10\,{\rm cm})^{-2}$.
& Static point charges in vacuum; fixed localized distributions; no screening, polarization, image forces, or external fields.
& M1: $r^{-2}\!\to r^{-1}$; M2: $p=1.20$; M3: $u_{q_1}u_{q_2}$ interaction.
& Sweep separation over its range, then independently cross both charges at a fixed separation. \\
\addlinespace[5pt]
\texttt{kepler\_\allowbreak orbit}
& Kepler, \textit{Harmonices mundi libri V} (1619) 
& $\mathcal{C}$: semi-major axis $a\in[0.2,8]$ AU; eccentricity $e\in[0,0.7]$. $\mathcal{R}$: orbital period (days).
& $P=365.25\,{\rm days}$\newline$\times(a/1\,{\rm AU})^{1.5}(1+e)^0$.
& Bound Newtonian two-body orbit about the same fixed central mass; negligible test mass, perturbations, tides, drag, and relativistic corrections.
& M1: $a^{1.5}\!\to a^{1}$; M2: $p=0.80$; M3: $u_au_e$ interaction.
& Sweep semi-major axis across more than one decade and vary eccentricity at a shared off-reference axis. \\
\addlinespace[5pt]
\texttt{brownian\_\allowbreak diffusion}
& Einstein, Brownian-motion paper (1905) 
& $\mathcal{C}$: lag $t\in[0.05,5]$ s; tracer radius $r\in[0.2,1]$ $\mu$m. $\mathcal{R}$: two-dimensional MSD ($\mu$m$^2$).
& $\mathrm{MSD}_{2D}=8\,{\rm \mu m^2}$\newline$\times(t/1\,{\rm s})^1$\newline$\times(r/0.5\,{\rm \mu m})^{-1}$.
& Overdamped dilute spheres in a Newtonian fluid; fixed temperature and viscosity; far from walls; lag below drift/confinement times.
& M1: $t^{1}\!\to t^{2}$; M2: $p=1.20$; M3: $u_tu_r$ interaction.
& Sweep lag over two decades at fixed radius, then cross lag and radius at separated levels. \\
\addlinespace[5pt]
\texttt{poiseuille\_\allowbreak flow}
& Hagen (1839) and Poiseuille (preliminary reports 1840--1841; memoir issued 1844 and collected 1846) 
& $\mathcal{C}$: $\Delta P\in[2,10]$ kPa; radius $r\in[0.25,0.5]$ mm; length $L\in[20,50]$ cm; viscosity $\mu\in[1,3]$ mPa s. $\mathcal{R}$: flow rate (mL s$^{-1}$).
& $Q=1.2271846303\,{\rm mL\,s^{-1}}$\newline$\times(\Delta P/10\,{\rm kPa})^1$\newline$\times(r/0.5\,{\rm mm})^4$\newline$\times(L/20\,{\rm cm})^{-1}$\newline$\times(\mu/1\,{\rm mPa\,s})^{-1}$.
& Steady fully developed incompressible Newtonian laminar flow; rigid straight circular tube; no-slip wall; losses measured only between pressure taps.
& M1: $r^{4}\!\to r^{3}$; M2: $p=0.80$; M3: $u_{\Delta P}u_r$ interaction.
& Sweep radius at fixed pressure, length, and viscosity; cross pressure and radius, then verify retained inverse length/viscosity factors. \\
\end{xltabular}
\endgroup

\begingroup
\scriptsize
\setlength{\tabcolsep}{2pt}
\renewcommand{\arraystretch}{1.15}
\setlength{\LTleft}{0pt}
\setlength{\LTright}{0pt}
\setlength{\LTcapwidth}{\textwidth}
\setcounter{LTchunksize}{200}
\begin{longtable}{@{}
>{\RaggedRight\arraybackslash}p{\dimexpr 0.09999999999999999\textwidth-2.3999999999999995pt\relax}
>{\RaggedRight\arraybackslash}p{\dimexpr 0.10714285714285714\textwidth-2.5714285714285716pt\relax}
>{\RaggedRight\arraybackslash}p{\dimexpr 0.17142857142857143\textwidth-4.114285714285714pt\relax}
>{\RaggedRight\arraybackslash}p{\dimexpr 0.21428571428571427\textwidth-5.142857142857143pt\relax}
>{\RaggedRight\arraybackslash}p{\dimexpr 0.15\textwidth-3.6000000000000005pt\relax}
>{\RaggedRight\arraybackslash}p{\dimexpr 0.12142857142857143\textwidth-2.914285714285714pt\relax}
>{\RaggedRight\arraybackslash}p{\dimexpr 0.1357142857142857\textwidth-3.2571428571428567pt\relax}@{}}
\caption{\textbf{The 24 families in the Temporal collection}. These are derived from 23 primary papers and paired with benchmark-authored counterfactual responses. The source-reported branch is a deterministic implementation of the selected relation under the stated approximations. It does not reproduce all experiments or raw data in the paper. Numerical intervals and discrete sets below are the current legal query domains.}\label{tab:temporal33}\\
\toprule
\textbf{Family} & \textbf{Journal and source} & \textbf{Experimental interface} & \textbf{Source-reported relation} & \textbf{Validity scope} & \textbf{Counterfactual construction} & \textbf{Separating queries} \\
\midrule
\endfirsthead
\multicolumn{7}{l}{\tablename~\thetable{} (continued)}\\
\toprule
\textbf{Family} & \textbf{Journal and source} & \textbf{Experimental interface} & \textbf{Source-reported relation} & \textbf{Validity scope} & \textbf{Counterfactual construction} & \textbf{Separating queries} \\
\midrule
\endhead
\midrule
\multicolumn{7}{r}{\textit{Continued on next page}}\\
\endfoot
\bottomrule
\endlastfoot

Defect supersymmetric entropy
& \textit{Physical Review Letters}\newline\cite{temporal2026_01}.
& $\mathcal C$ inverse R\'{e}nyi index $\gamma\in[0.5,\allowbreak 2]$. $\mathcal R$ normalized entropy $y$ (dimensionless).
& $y=12[1+0.125(\gamma-1)]$.
& Symmetric $k=2$ Wilson surface in the $A_3$ $(2,\allowbreak 0)$ theory. $r_1=r_2=1/2$. Exact relation at fixed theory and representation.
& M1 replaces $\gamma$ by $\gamma^2$, preserving $y(1)=12$.
& Query $\gamma=0.5$ and $2$, and retain $\gamma=1$ as a common reference. \\
\addlinespace[5pt]

Turbulent scaling exponent
& \textit{Physical Review Letters}\newline\cite{temporal2026_02}.
& $\mathcal C$ moment order $p\in[1.4,\allowbreak 4.6]$. $\mathcal R$ scaling exponent $\zeta_p$ (dimensionless).
& $\zeta_p=p/3-Cp(p-2)\epsilon^2$,\newline $C=(2^{2/3}+1)/(4\,\allowbreak 2^{1/3}\ln2)$ and $\epsilon=0.1$.
& Inertial-range shell-model asymptotics through $O(\epsilon^2)$. Terms $O(\epsilon^4)$ are omitted. Continuous moment orders are benchmark queries.
& M1 changes $\epsilon$ from $0.1$ to $0.2$. The correction is multiplied by four and $\zeta_2=2/3$ is retained.
& Sample moment orders below and above $p=2$, including $1.4$ and $4.6$. The protected $p=2$ alone does not separate the laws. \\
\addlinespace[5pt]

MnF$_2$ crossover displacement
& \textit{Physical Review Letters}\newline\cite{temporal2026_03}.
& $\mathcal C$ strain $\epsilon_{xy}\in\{-0.0004,\allowbreak -0.0008,\allowbreak -0.0011,\allowbreak -0.0015\}$ and field $B\in[0.644,\allowbreak 5.796]$ T. $\mathcal R$ crossover shift $\Delta T^*$ (K).
& $T_c=67.467-59\epsilon_{xy}-0.01585B^2$.\newline $\Delta T^*=T_c[0.56(0.67171381472)\cdot(-B\epsilon_{xy})/T_c]^{2/3}$.
& Compression along $[110]$, field along $c$. Mean-field exponent $2/3$. All admitted strains are negative, so $-B\epsilon_{xy}=|B\epsilon_{xy}|$.
& M1 replaces every $B$ by $W_{1-\epsilon_{xy}/0.0015}(B)$, including its occurrences in $T_c$.
& Cross strain levels with interior fields, including $B=3.22$ T. Endpoint fields are preserved by the composition. \\
\addlinespace[5pt]

Confined ciliate swimming
& \textit{Physical Review Letters}\newline\cite{temporal2026_04}.
& $\mathcal C$ mean slip $u\in[-0.4,\allowbreak 0.4]$ and amplitude $a\in[0.24,\allowbreak 1.36]$. $\mathcal R$ normalized laboratory-frame mean swimming velocity $\langle v_p\rangle/v_m$. All dimensionless.
& $\langle v_p\rangle/v_m=\sqrt{\max[(1-u)^2-a^2,\allowbreak 0]}-1$.
& Resonant channel and metachronal wavelengths, lubrication and incompressible Stokes flow. Fixed reduced mobility $|G_n|=2$. Includes the phase-locked plateau.
& M1 and M2 replace $u$ and $a$ by $W_2(u)$ and $W_2(a)$.
& Cross interior slip and amplitude settings through the oscillatory-to-locked transition. Include interior values because corners alone are preserved. \\
\addlinespace[5pt]

Quantum interconnect rate
& \textit{Physical Review Letters}\newline\cite{temporal2026_05}.
& $\mathcal C$ modes $M\in\{2,\allowbreak 4,\allowbreak 8,\allowbreak 11\}$, storage efficiency $\eta_s\in[0.54,\allowbreak 0.86]$ and noise $p_n\in\{0.01,\allowbreak 0.10,\allowbreak 0.50\}$. $\mathcal R$ scaled rate $\widetilde R=(L_0/c)R_{\rm qm}$ (bits).
& $A=1-(M-1)p_n/M$.\newline $\mathcal I=\log_2M+A\log_2A+(M-1)(p_n/M)\log_2(p_n/M)$.\newline $\widetilde R=0.7e^{-250/44}\mathcal I\eta_s^4/[27(2-\eta_s)(4-3\eta_s)]$.
& Ideal 1000-km, four-segment repeater with two nesting levels. $N=\eta_d=1$, $L_{\rm att}=22$ km. Equal input probabilities and depolarizing memory noise. Lifetime losses omitted.
& M1 replaces all $\eta_s$ occurrences by $W_{1+p_n/0.1}(\eta_s)$, retaining the same information term.
& Cross noise and mode levels at interior $\eta_s=0.7$. Efficiency endpoints alone are preserved. \\
\addlinespace[5pt]

Time-dependent CP asymmetry
& \textit{Physical Review Letters}\newline\cite{temporal2026_07}.
& $\mathcal C$ proper time $t\in[1.25,\allowbreak 7.25]$ ps. $\mathcal R$ integrated CP asymmetry $A_{CP}$ (dimensionless).
& $A_{CP}(t)=0.82\sin(0.507t)+0.13\cos(0.507t)$.
& Helicity-angle-integrated $B^0\to K_S^0\mu^+\mu^-$ asymmetry. $\Delta\Gamma_d=0$ approximation and fixed dilepton-mass exclusions. Proper time is an observation coordinate.
& M1 replaces both time occurrences by $W_2(t)$.
& Sample interior proper times across the oscillation, in addition to $1.25$ and $7.25$ ps. The two endpoints alone coincide. \\
\addlinespace[5pt]

Fermentation minimum selling price
& \textit{Nature Communications}\newline\cite{temporal2026_08}.
& $\mathcal C$ yield $Y\in[0.28,\allowbreak 0.92]$ g product/g sugar and titer $T\in[21,\allowbreak 109]$ g product/L. $\mathcal R$ minimum selling price $P$ (2019 USD/kg product).
& $P=[0.311+0.353/Y+10.9/T+0.356/(YT)]/0.983$.
& Fitted techno-economic relation for glucose-to-succinic acid, low-pH index 25. Productivity $0.66$ g L$^{-1}$ h$^{-1}$ and recovery $0.983$. Large-scale fed-batch plant model.
& M1 and M2 replace yield $Y$ and titer $T$ by $W_2(Y)$ and $W_2(T)$ in every occurrence.
& Cross interior yield and titer levels with their endpoints to distinguish both reciprocal terms and their interaction. \\
\addlinespace[5pt]

Shear-thinning cutting force
& \textit{Nature Communications}\newline\cite{temporal2026_13}.
& $\mathcal C$ $r\in\{0.6856,\allowbreak 0.9775,\allowbreak 1.0829,\allowbreak 1.4455,\allowbreak 1.5566\}$ $\mu$m. $v\in\{0.083,\allowbreak 0.125,\allowbreak 0.167,\allowbreak 0.208,\allowbreak 0.250,\allowbreak 0.292,\allowbreak 0.333,\allowbreak 0.375,\allowbreak 0.417,\allowbreak 0.458,\allowbreak 0.500\}$ m/s. $h\in\{5,\allowbreak 23,\allowbreak 33\}$ mm. $\mathcal R$ relative force $y=F/F_{\rm ref}$ (dimensionless).
& $y=(r/0.6856)^{0.968}\cdot(v/0.5)^{0.72}\cdot(h/33)^{1.244}$.
& Printed empirical fit for the fixed optimized starch fluid and calibrated straight blades. Discrete joint query grid. Other compositions, blade geometries and the unlabelled 16-mm-sheet series are excluded.
& M1 replaces the radius exponent $0.968$ by $0.968(v/0.5)$, coupling radius and speed. The common normalization is retained.
& Cross at least two nonreference radii with low and high speeds, with depth fixed. The reference radius or reference speed alone hides the mutation. \\
\addlinespace[5pt]

Off-diagonal scattering density
& \textit{Physical Review Letters}\newline\cite{temporal2026_14}.
& $\mathcal C$ cross section $\sigma\in[0.04,\allowbreak 0.36]$. $\mathcal R$ local probability-density value $p_\sigma$. Both coordinates are dimensionless.
& $m=0.967^2/(2(0.967)+7.013)$, $z=\sigma/m$,\newline $a=(2(0.967)-1)/(2(0.967)+7.013)$.\newline $p_\sigma=e^{-z}[1+a((4-z)z-2)]/m$.
& Leading and first subleading Heidelberg asymptotic density for $\beta=2$, $T_1=T_2=0.967$ and absorption $7.013$. Used near the Ericson transition. The finite query interval is not a separately normalized support.
& M1 composes $p_\sigma$ with $W_2(\sigma)$. M2 multiplies it by $W_2^{\prime}(\sigma)=2(\sigma-0.04)/0.32$.
& Query endpoints and interior cross sections. The Jacobian changes density heights even where the coordinate composition fixes an endpoint. \\
\addlinespace[5pt]

Thin-film differential reflectance
& \textit{Physical Review Letters}\newline\cite{temporal2026_15}.
& $\mathcal C$ SnS thickness $d\in[72.05,\allowbreak 252.45]$ nm. $\mathcal R$ AC-polarized differential reflectance $\Delta R/R$ (dimensionless).
& $R=|(r_{01}+r_{12}z)/(1+r_{01}r_{12}z)|^2$,\newline $z=e^{4\pi i\widetilde n d/800}$, $r_{01}=(1-\widetilde n)/(1+\widetilde n)$, $r_{12}=(\widetilde n-1.7276)/(\widetilde n+1.7276)$.\newline $\Delta R/R=-0.00302\,\allowbreak \partial_n\ln R$ at $\widetilde n=n+0.12i$, $n=4.23$.
& Normal-incidence 800-nm light in air/SnS/MgO. First-order real-index perturbation only, with imaginary-index change omitted. All multiple reflections and the analytic derivative are retained.
& M1 replaces every thickness occurrence by $W_2(d)$ in the complete differentiated Fresnel response.
& Sweep interior thickness through interference extrema. The original thickness endpoints are common to both laws. \\
\addlinespace[5pt]

Squeezed-vacuum energy surface
& \textit{Physical Review Letters}\newline\cite{temporal2026_16}.
& $\mathcal C$ bosonic coordinates $q,\allowbreak p\in[-1.6,\allowbreak 1.6]$. $\mathcal R$ thermodynamic-limit energy $h_\infty$ in the fixed model energy unit.
& $h_\infty=(q^2+p^2)/2-2\sqrt{1+0.04q^2(q^2+p^2)}$.
& Mean-field squeezed-vacuum surface after eliminating atomic coordinates. $\omega=1$, $\omega_0=2$, $\gamma/\omega=0.4$, below spectral collapse. Not an exact finite-system energy.
& M1 and M2 replace all $q$ and $p$ occurrences by $W_2(q)$ and $W_2(p)$.
& Cross interior positive, zero and negative phase-space coordinates to test curvature and the source reflection symmetries. \\
\addlinespace[5pt]

Pulsed magnetic-field sensing
& \textit{Physical Review X}\newline\cite{temporal2026_17}.
& $\mathcal C$ field $B\in[10^{-8},\allowbreak 9\!\times\!10^{-8}]$ T and pulse blocks $N\in\{10,\allowbreak 30,\allowbreak 58,\allowbreak 80,\allowbreak 100\}$. $\mathcal R$ signal-to-noise ratio.
& $t_c=1.78\!\times\!10^{-6}N$ s.\newline $A_N=\sqrt{60/(t_c+0.004)}$\newline $\times e^{-t_c/0.000104}/\mathcal N$.\newline $\mathrm{SNR}=A_N\sin(0.34\gamma_eBt_c)$,\newline $\gamma_e=176086268505.04355$ s$^{-1}$ T$^{-1}$ and $\mathcal N\simeq14.109773394162719$.
& Fixed acquisition and noise parameters of the nonlinear sensitivity model. Sinusoidal field at a filter-function lobe and sensing duration much shorter than $T_1$. $\mathcal N$ is evaluated from the frozen source noise expression.
& M1 replaces $B$ only inside the phase by $W_{1+N/100}(B)$. Acquisition, decay and noise terms retain their forms.
& Cross interior field strengths with distinct block counts. Field endpoints alone conceal the composition. \\
\addlinespace[5pt]

Undetected-light visibility
& \textit{Nature Communications}\newline\cite{temporal2026_18}.\newline Advance publication.
& $\mathcal C$ object-arm intensity transmission $T\in[0.1,\allowbreak 0.9]$. $\mathcal R$ fringe visibility $V$. Both dimensionless.
& $a=0.784^2(0.47)T$.\newline $b=0.8(0.118)$ and $c=0.885^2+0.8$.\newline $V=\dfrac{2\sqrt{0.8a}(0.885+\sqrt b)}{b+c(1+a)}$.
& Michelson nonlinear thin-film model with undepleted pump and seed. Stimulated emission dominates. Only the first four-wave-mixing reflection and within-window interference are retained.
& M1 replaces both transmission occurrences by $W_2(T)$.
& Scan interior transmission values between 0.1 and 0.9. The two endpoints alone agree. \\
\addlinespace[5pt]

Davemaoite phase boundary
& \textit{Nature Communications}\newline\cite{temporal2026_19}.
& $\mathcal C$ pressure $P\in[36,\allowbreak 124]$ GPa. $\mathcal R$ transition temperature $T_{\rm tr}$ (K).
& $T_{\rm tr}=533+2.4P$.
& Printed fitted tetragonal-to-cubic phase boundary for lower-mantle davemaoite. Deterministic use of central fit coefficients. Fit uncertainties are not per-query noise.
& M1 replaces $P$ by $W_2(P)$.
& Compare at least two interior pressures against the source straight line, with endpoints as shared references. \\
\addlinespace[5pt]

CoO orbital magnetoresistance
& \textit{Science}\newline\cite{temporal2026_20}.
& $\mathcal C$ remaining Cu* thickness $t\in[3.88,\allowbreak 5.32]$ nm. $\mathcal R$ orbital-magnetoresistance amplitude $A$ (percent).
& $A=0.32+0t$ percent.
& Approximate plateau of CoO(5 nm)/Cu* at 150 K. One Hall bar is progressively ion-milled and exposed to air for 12 h after each etch. Thickness is resistance-calibrated.
& M1 changes the affine slope from $0$ to $0.08$ percent/nm, giving $A=0.32+0.08t$.
& Compare low and high remaining thickness to test the plateau against a nonzero slope under the fixed protocol. \\
\addlinespace[5pt]

MnF$_2$ transition baseline
& \textit{Physical Review Letters}\newline\cite{temporal2026_03}.
& $\mathcal C$ loading-path strain $\epsilon\in\{-0.0015,\allowbreak -0.0011,\allowbreak -0.0008,\allowbreak -0.0004\}$ and field $B\in\{0,\allowbreak 6.44\}$ T. $\mathcal R$ baseline $T_c$ (K).
& $T_c=67.467-59\epsilon-0.01585B^2$.
& Fitted non-altermagnetic baseline along compressive $[110]$ loading, including co-induced symmetric strain. Not independently imposed pure shear or the shifted crossover $T^*$.
& M1 replaces $B^2$ by $B^4/B_0^2$, with fixed $B_0=1$ T. The field coefficient retains units K/T$^2$.
& Compare $B=0$ and $6.44$ T at matched strain, then repeat across strain levels. \\
\addlinespace[5pt]

Quasiperiodic localization exponent
& \textit{Physical Review Letters}\newline\cite{temporal2026_22}.
& $\mathcal C$ comprises couplings $\lambda_1,\allowbreak \lambda_2\in[0.20,\allowbreak 0.67]$. $\mathcal R$ is the Lyapunov exponent $L$. All quantities are dimensionless.
& $\Lambda=\dfrac{\lambda_2(1+\sqrt{1-\lambda_1^2})}{\lambda_1(1+\sqrt{1-\lambda_2^2})}$.\newline $L=\max(0,\allowbreak \ln\Lambda)$.
& The source-theory unitary almost-Mathieu relation is evaluated at non-Hermiticity $\eta=0$, irrational quasiperiodic field $\Phi=(\sqrt5-1)/2$ and phase $\theta=0$. Positive $L$ gives inverse localization length. A zero value covers extended or critical states. The interface returns the asymptotic theoretical exponent without finite-time photon-counting fluctuations.
& M1 and M2 replace every occurrence of $\lambda_1$ and $\lambda_2$ by $W_2(\lambda_1)$ and $W_2(\lambda_2)$.
& Cross distinct interior couplings with $\lambda_2>\lambda_1$, including $(0.25,\allowbreak 0.50)$. Include the reversed pair as a zero-response control. Equal couplings and domain corners alone do not separate the branches. \\
\addlinespace[5pt]

Viscous--inertial suspension rheology
& \textit{Physical Review Letters}\newline\cite{temporal2026_23}.
& $\mathcal C$ is the Stokes number $\mathrm{St}\in[0.01,\allowbreak 300]$. $\mathcal R$ is normalized macroscopic friction $r=\mu/\mu_c$. Both are dimensionless.
& $r=1+14.3\sqrt{\dfrac{0.065(1+0.029\,\allowbreak \mathrm{St})}{1+0.1\,\allowbreak \mathrm{St}}}$.
& The source-reported dense-suspension relation is restricted to interparticle friction $\mu_f=0.15$ and the constructed packing-related slice $K_\phi=J+0.1I^2=0.065$, with $\mathrm{St}=I^2/J$. The displayed coefficients specify the fixed deterministic response. Changing Stokes number does not independently scan packing or particle friction, and the admitted interval is not extrapolated.
& M1 replaces both Stokes-number occurrences by $W_2(\mathrm{St})$.
& Sweep logarithmically separated interior values such as $0.1$, $10$ and $100$. Compare the crossover curvature and use the shared endpoints only as controls. \\
\addlinespace[5pt]

Splashing-drop rim response
& \textit{Physical Review Letters}\newline\cite{temporal2026_25}.
& $\mathcal C$ is $\mathrm{We}\in[288,\allowbreak 592]$. $\mathcal R$ comprises rim speed $v_t=V_t/V$ and rim diameter $h_t=H_t/R_d$. All quantities are dimensionless.
& $\mathrm{Re}=\sqrt{\mathrm{We}}/0.0105$, $A=\sqrt3/(2\mathrm{Re})$, $b=1/\mathrm{We}$.\newline Let $x>0$ solve $x^4-bx-A=0$ and set $t_e=x^2$.\newline $v_t=0.45\sqrt{3/t_e}$, $h_t=2.2t_e^{3/2}$.
& The calibrated low-viscosity, high-Reynolds-number approximation describes the early two-jets spray regime. The fixed layer depth is $H/R_d=1.8$, Ohnesorge number $0.0105$, and gas density and viscosity ratios $0.001$ and $0.018$. Both readouts describe the toroidal rim at $t_0=0.07$ in units of $R_d/V$, including ballistic rim motion and early thickening.
& M1 replaces every Weber-number occurrence by $W_2(\mathrm{We})$ in the linked root construction and both outputs.
& Sample several interior Weber numbers and compare the coupled speed and diameter curves. Preserve the low and high endpoints as common references rather than using them alone. \\
\addlinespace[5pt]

Trapped-ion optical Magnus response
& \textit{Physical Review Letters}\newline\cite{temporal2026_26}.
& $\mathcal C$ comprises focal-plane positions $x,\allowbreak y\in[-1.04,\allowbreak 1.04]$ $\mu$m and transition category $q\in\{-1,\allowbreak +1\}$. $\mathcal R$ is ground-state survival probability $P_q$, dimensionless.
& $g_q=(x/w)(1+qy/z_R)e^{-(x^2+y^2)/w^2}$, $w=1.3\,\allowbreak \mu\mathrm m$, $z_R=\pi w^2/(0.729\,\allowbreak \mu\mathrm m)$.\newline $a=\sqrt{1+2(w/z_R)^2}$, $g_*=\dfrac{a+1}{2\sqrt{2e}}e^{-(w/z_R)^2/(a+1)^2}$.\newline $P_q=\cos^2[\pi|g_q|/(2g_*)]$.
& A resonant single pulse addresses a fixed $^{40}\mathrm{Ca}^+$ ion initially in $m_g=+1/2$. The beam propagates along $z$, is polarized along $y$, and the quantization field points along $-x$. The first correction beyond the paraxial quadrupole-gradient model is retained. Each transition uses its modeled peak coupling over the one-waist square for its own fixed $\pi$-pulse calibration.
& M1 replaces both occurrences of $x$ by $W_2(x)$. The coordinate $y$, discrete transition category and fixed pulse calibration are retained.
& Cross interior horizontal positions, especially $x=0$, with positive and negative vertical positions and both transition categories. Track the survival-profile displacement while retaining the horizontal endpoints as controls. \\
\addlinespace[5pt]

Parity-prepared oscillator squeezing
& \textit{npj Quantum Information}\newline\cite{temporal2026_27}.\newline Advance publication.
& $\mathcal C$ comprises coherent-displacement amplitude $a\in[0.14,\allowbreak 1.26]$ and per-step intensity transmissivity $\eta\in[0.865,\allowbreak 0.985]$, both dimensionless. $\mathcal R$ is final conditional squeezing $S$ in dB.
& $\Pi_\beta=D(\beta)[1+(-1)^{\hat n}]D^\dagger(\beta)/2$, $\mathcal M_\beta(\rho)=\mathcal L_\eta(\Pi_\beta\rho\Pi_\beta)$ with full pure loss $\mathcal L_\eta$.\newline Normalize $\rho_*\propto\mathcal M_0\mathcal M_{-ia}\mathcal M_{ia}(|0\rangle\langle0|)$.\newline $S=-10\log_{10}[4\min_\theta\operatorname{Var}_{\rho_*}(x_\theta)]$, $x_\theta=(\hat a e^{-i\theta}+\hat a^\dagger e^{i\theta})/2$. The card gives the exact nine-exponential reduction.
& Three ideal displaced-parity measurements start from vacuum and retain only the all-even outcomes. Full unobserved pure loss follows each measurement, including the third, and the final state is normalized by its full success probability. The response retains coherences and all loss events. The displacement amplitude is not time, and no Gaussian-state or single-loss approximation is used.
& M1 replaces all 19 amplitude occurrences by $a^2/1.4$. This uses the preregistered source amplitude reference, not $W_2$ on the smaller public interval.
& Cross low, middle and high admitted amplitudes with several transmissivities. Compare the complete squeezing curve. The public amplitude endpoints are not fixed by this composition. \\
\addlinespace[5pt]

Mean-field microcanonical coexistence
& \textit{Physical Review X}\newline\cite{temporal2026_28}.
& $\mathcal C$ comprises $h_x\in[0.06,\allowbreak 0.14]$ and $h_z\in[-0.24,\allowbreak -0.16]$. $\mathcal R$ comprises crossing energy density $\epsilon_\times$, common magnetization magnitude $m_\times$ and signed branches $m_z^-,\allowbreak m_z^+$. All are dimensionless in the stated Pauli and $J_3$ units.
& $s=-\sqrt{(-h_z+\sqrt3h_x)/3}$.\newline $\epsilon_\times=2s^3$, $m_\times=\dfrac23\sqrt{-3h_z+\sqrt3h_x}$.\newline $m_z^\pm=\dfrac{s\pm\sqrt{s^2-8h_z/3}}2$.
& The uniform thermodynamic LMG-3 mean-field problem maximizes microcanonical entropy at fixed energy $\epsilon=z^3+h_zz+h_xx$ over $x^2+z^2\le1$. The selected negative-energy crossing has two global entropy maxima with opposite signed longitudinal magnetizations and the same magnitude. The Pauli magnetization convention is used. This target excludes finite-size disorder, tunneling rates and dynamical ETH hybridization thresholds.
& M1 replaces every $h_x$ occurrence in all four outputs by $0.05+(h_x-0.05)^2/0.10$. The fixed references are the original safety-box endpoints $0.05$ and $0.15$.
& Cross transverse and longitudinal fields over the legal box. Compare the crossing energy and both signed branches together. The current public transverse-field endpoints do not remain invariant. \\
\addlinespace[5pt]

Loss-mediated transfer and entanglement
& \textit{Physical Review A}\newline\cite{temporal2026_31}.
& $\mathcal C$ is scaled time $\tau=\kappa t\in[30,\allowbreak 270]$. $\mathcal R$ comprises $P_{L|L},\allowbreak P_{R|L},\allowbreak C_L,\allowbreak P_{L|R},\allowbreak P_{R|R},\allowbreak C_R$, all dimensionless. The suffix specifies the independent initial qubit excitation, and $C_d$ is reduced two-qubit concurrence.
& $u=2^{-1/4}/10$, $\mathbf a^{(d)}=e^{-iH_{\rm eff}\tau}\mathbf e_d$.\newline $H_{\rm eff}=\left(\begin{smallmatrix}
0&0&u&u\\
0&0&u&iu\\
u&u&-\frac{1+i}{2}&0\\
u&-iu&0&\frac{1-i}{2}
\end{smallmatrix}\right)$ in basis $(L,\allowbreak R,\allowbreak c_1,\allowbreak c_2)$.\newline $P_{j|d}=|a_j^{(d)}|^2$, $C_d=2|a_L^{(d)}a_R^{(d)}|$. Amplitudes are unnormalized and lost probability occupies the vacuum.
& The engineered sideband-selected rotating-wave model retains both equal-loss cavities with detunings $\Delta_{1,\allowbreak 2}/\kappa=\pm1/2$ and relative coupling phase $-\pi/2$. Each reset starts with one specified qubit excited and both cavities in vacuum. Cavity recycling is unconditional. Intrinsic qubit decoherence and thermal excitation are omitted. No cavity elimination or exactly vanishing reverse transfer is imposed.
& M1 replaces all 48 time occurrences in the six complete modal outputs by $W_2(\tau)$. The populations and concurrence require no derivative factor.
& Sample several interior times from both preparations, including early and late transients. Compare transferred populations and concurrence jointly. The interval endpoints remain shared. \\
\addlinespace[5pt]

Non-Markovian cooperative radiation
& \textit{Quantum}\newline\cite{temporal2026_32}.
& $\mathcal C$ is scaled time $\tau=\gamma_0t\in[2,\allowbreak 18]$. $\mathcal R$ comprises unconditional Dicke populations $P_2,\allowbreak P_1,\allowbreak P_0$ and signed atomic energy-loss rate $j=I/(\omega_0\gamma_0)$. All four are dimensionless.
& $\partial_\tau\rho=-i[H,\allowbreak \rho]+\mathcal D[b]\rho$, $H=(J_-b^\dagger+J_+b)/\sqrt2$, $\rho(0)=|ee,\allowbreak 0\rangle\langle ee,\allowbreak 0|$.\newline $\mathcal D[b]\rho=b\rho b^\dagger-\{b^\dagger b,\allowbreak \rho\}/2$.\newline $P_n=\langle n|\operatorname{Tr}_b\rho|n\rangle$ for symmetric Dicke states $n=0,\allowbreak 1,\allowbreak 2$.\newline $j=-\partial_\tau(2P_2+P_1)$. The card retains every modal coefficient.
& Two identical atoms have common coupling phase, a vacuum Lorentzian reservoir, linewidth $\lambda/\gamma_0=1/2$ and transition ratio $\omega_0/\gamma_0=1000$. The resonant rotating-wave and narrowband pseudomode model retains all reachable excitation sectors and loss recycling. The initial atomic state is fully excited. Negative signed intensity represents reabsorption, and the three populations sum to one.
& M1 composes all populations and signed intensity with $g(\tau)=W_2(\tau)$. M2 multiplies the composed intensity by $g'(\tau)=(\tau-2)/8$, preserving its derivative definition.
& Sample interior early and late times and compare populations with signed energy flow. Endpoint populations agree between branches, whereas endpoint intensities need not because of the derivative factor. \\
\end{longtable}
\endgroup

\end{document}